\documentclass[sigconf]{acmart}

\AtBeginDocument{%
  }

\usepackage[table]{xcolor} 
\usepackage{multirow}

\usepackage[ruled,vlined]{algorithm2e}
\DontPrintSemicolon
\SetKwInput{Input}{Input}
\SetKwInput{Output}{Output}

\usepackage{hyperref}

\acmSubmissionID{59}

\begin{document}

\title{OmniEarth: A Benchmark for Evaluating Vision–Language Models in Geospatial Tasks}

\author{Ronghao Fu}
\authornote{Both authors contributed equally to this research.}
\author{Haoran Liu}
\authornotemark[1]
\affiliation{%
  \institution{Jilin University}
  \city{Changchun}
  \country{China}
}
\email{furh@jlu.edu.cn, lhr24@mails.jlu.edu.cn}

\author{Weijie Zhang}
\affiliation{%
  \institution{Jilin University}
  \city{Changchun}
  \country{China}
}
\email{weijie25@mails.jlu.edu.cn}

\author{Zhiwen Lin}
\affiliation{%
  \institution{Jilin University}
  \city{Changchun}
  \country{China}
}
\email{linzw25@mails.jlu.edu.cn}

\author{Xiao Yang}
\affiliation{%
  \institution{Jilin University}
  \city{Changchun}
  \country{China}
}
\email{yangx23@jlu.edu.com}

\author{Peng Zhang}
\affiliation{%
  \institution{Chang Guang Satellite Technology}
  \city{Changchun}
  \country{China}}
\email{zhangpeng1@jl1.cn}

\author{Bo Yang}
\authornote{Corresponding author.}
\affiliation{%
  \institution{Jilin University}
  \city{Changchun}
  \country{China}}
\email{ybo@jlu.edu.cn}

\renewcommand{\shortauthors}{Fu et al.}

\begin{abstract}
Vision–Language Models (VLMs) have demonstrated effective perception and reasoning capabilities on general-domain tasks, leading to growing interest in their application to Earth observation. However, a systematic benchmark for comprehensively evaluating remote sensing vision–language models (RSVLMs) remains lacking.
To address this gap, we introduce OmniEarth, a benchmark for evaluating RSVLMs under realistic Earth observation scenarios. OmniEarth organizes tasks along three capability dimensions: perception, reasoning, and robustness. It defines 28 fine-grained tasks covering multi-source sensing data and diverse geospatial contexts. The benchmark supports two task formulations: multiple-choice VQA and open-ended VQA. The latter includes pure text outputs for captioning tasks, bounding box outputs for visual grounding tasks, and mask outputs for segmentation tasks. To reduce linguistic bias and examine whether model predictions rely on visual evidence, OmniEarth adopts a blind test protocol and a quintuple semantic consistency requirement. OmniEarth includes 9,275 carefully quality-controlled images, including proprietary satellite imagery from Jilin-1 (JL-1), along with 44,210 manually verified instructions. 
We conduct a systematic evaluation of contrastive learning-based models, general closed-source and open-source VLMs, as well as RSVLMs. Results show that existing VLMs still struggle with geospatially complex tasks, revealing clear gaps that need to be addressed for remote sensing applications. OmniEarth is publicly available at  \href{https://huggingface.co/datasets/sjeeudd/OmniEarth}{https://huggingface.co/datasets/sjeeudd/OmniEarth}.

\end{abstract}


\received{20 February 2007}
\received[revised]{12 March 2009}
\received[accepted]{5 June 2009}

\maketitle

\section{Introduction}

In recent years, Vision–Language Models (VLMs)~\cite{wang2025dynamic,zhang2024vision,wei2024vary,wang2022omnivl,shi2026iir} have demonstrated significant capabilities in multimodal perception and reasoning. Representative models, such as Qwen2.5-VL~\cite{Qwen2.5-VL}, GLM-V~\cite{glm}, and InternVL3.5~\cite{wang2025internvl3_5}, have facilitated progress in visual understanding, grounded reasoning, and natural language interaction by enabling the joint interpretation of visual and textual information. 

Building on this progress, the remote sensing (RS) community has increasingly explored Remote Sensing Vision–Language Models (RSVLMs), such as RemoteCLIP~\cite{liu2024remoteclip}, GeoChat~\cite{kuckreja2024geochat}, and LHRS-Bot~\cite{muhtar2024lhrs}. By fine-tuning general-purpose VLMs on large-scale RS image–text datasets, these models incorporate domain-specific geospatial information, extending their utility to tasks such as urban planning~\cite{PlanGPT-VL}, environmental assessment~\cite{10.1145/3764917.3771334}, and disaster management~\cite{wang2025disasterm3}. Despite this progress, applying VLMs to geospatial domains remains challenging. Satellite imagery exhibits large variations in object scale~\cite{10510173}, spatial resolution~\cite{10542972}, illumination ~\cite{wang2025illume}, and sensing modality~\cite{ni2022cross}, which challenge models primarily trained on natural images~\cite{lu2025vision}. In addition, many geospatial tasks require temporal analysis~\cite{sheikh2025application} to capture long-term changes, such as land-use change~\cite{ouyang2025object} or urban expansion~\cite{UrBench}, further increasing task complexity. The lack of benchmarks specifically designed for geospatial scenarios also limits objective evaluation, making it difficult to clearly identify model limitations and remaining gaps.

\begin{figure}[!t]
    \centering
    \includegraphics[width=\linewidth]{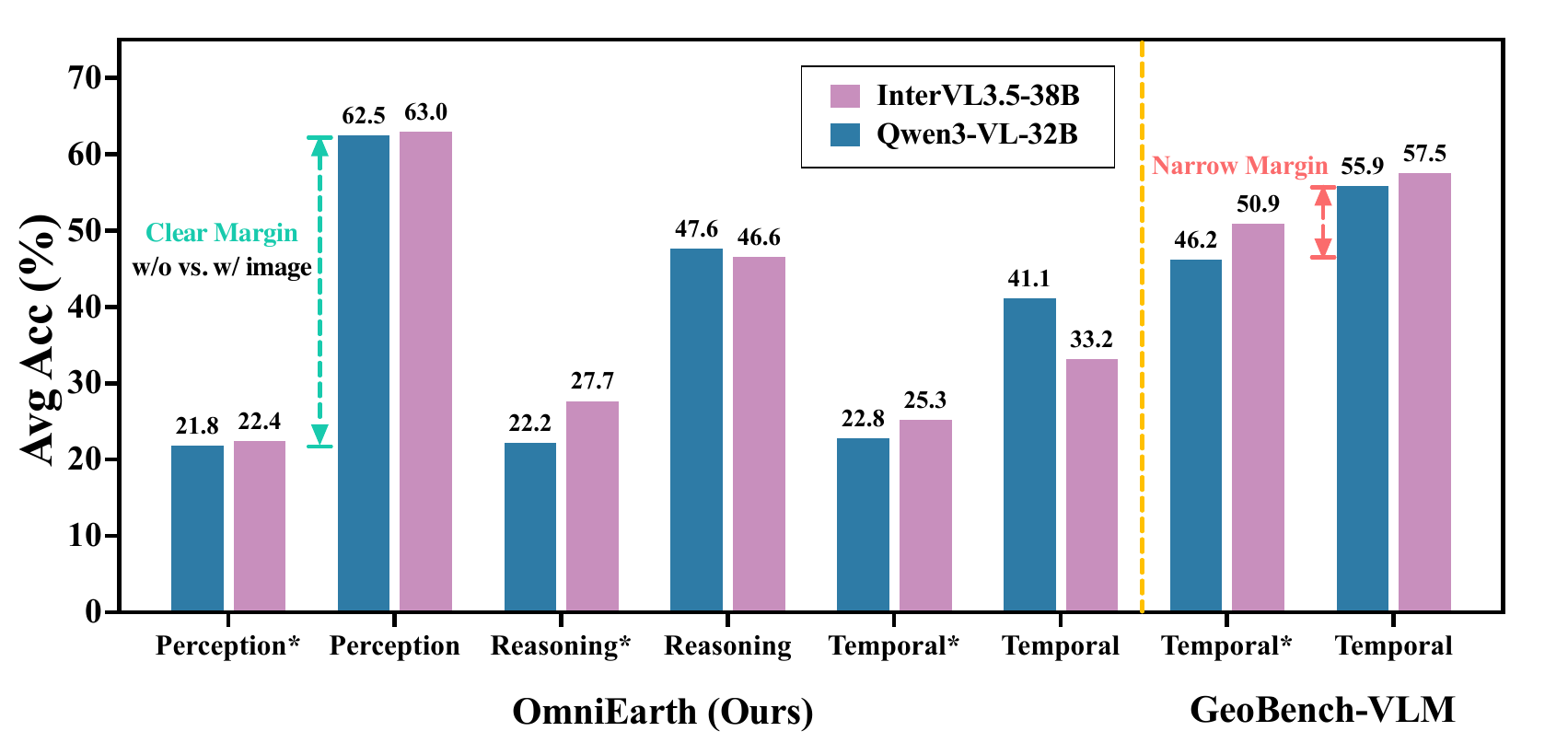}
    \caption{Blind Test Comparison Between Image-Based and Text-Only Settings. * denotes text-only evaluation without image input. Larger performance gaps indicate stronger reliance on visual evidence rather than language priors.}
    \label{fig:Preliminaryblind}
\end{figure}

To address evaluation needs in Earth observation, several benchmarks have been introduced, including VRSBench~\cite{li2024vrsbench}, DynamicVL~\cite{DynamicVL}, XLRSBench~\cite{wang2025xlrs}, CHOICE~\cite{CHOICE}, and GEOBench-VLM~\cite{danish2025geobenchvlm}. While these benchmarks provide useful testbeds for RSVLMs, they still have some limitations.
\textbf{First, existing benchmarks provide limited task granularity and coverage.} For example, CHOICE~\cite{CHOICE} does not include pixel-level analysis, implicit reasoning tasks, multi-temporal image analysis, or scenarios related to natural disasters, and offers limited evaluation under realistic imaging degradations. In addition, multimodal evaluation is often limited to image modality recognition, with few concrete tasks designed for other RS modalities such as SAR, and limited assessment of semantic alignment across modalities. GEOBench-VLM~\cite{danish2025geobenchvlm} targets Earth observation tasks, but its task taxonomy is relatively coarse. From a capability evaluation perspective, some tasks are mainly differentiated by object categories, such as subdividing object counting into vehicle counting and aircraft counting, while the core perceptual and reasoning capabilities required by these tasks are largely the same. Such task decomposition based on semantic categories increases the number of tasks to some extent, but does not substantially expand the range of model capabilities being evaluated.
\textbf{Second, existing benchmarks provide limited verification of whether model reasoning truly relies on visual information.} Most benchmarks~\cite{li2024vrsbench, danish2025geobenchvlm, DynamicVL} adopt a single question–answer format, and the answer options often lack sufficient discrimination, making it difficult to distinguish whether a model reasons based on image content or mainly relies on language priors. Preliminary experiments shown in Figure~\ref{fig:Preliminaryblind} indicate that, for some tasks, models achieve performance comparable to joint image–text input even when only text is provided and no image information is used. This indicates that current benchmarks have limited ability to separate visual understanding from language priors. In some cases, models can even select the correct answer by relying solely on the textual options through simple elimination, which fails to verify genuine multimodal analysis and reasoning. 
\textbf{Third, evaluation results may be affected by bias introduced through data reuse.} Many existing benchmarks are built on public remote sensing datasets, and some benchmarks directly reuse data from earlier benchmarks when constructing tasks. Since these datasets are often included in the pre-training of foundation models, such data overlap can, to some extent, reduce the reliability of zero-shot evaluation and make the assessment of model generalization less clear. In addition, some benchmarks~\cite{CHOICE} reuse the same images across different tasks, such as attribute recognition and spatial relationship recognition, which further reduces the diversity of the evaluation set.

\begin{table*}[!t]
\centering
\caption{Comprehensive Comparison of Remote Sensing Datasets and Benchmarks.}
\label{tab:comprehensive-rs-benchmarks}
\vspace{-0.35cm}
\footnotesize
\resizebox{\linewidth}{!}{
\begin{tabular}{llllccccc}
\toprule
\textbf{Benchmark/Dataset} & \textbf{Modalities} & \textbf{Data Sources} & \textbf{Geospatial Coverage} & \textbf{Answer Type} & \textbf{Reasoning} & \textbf{Human Verify} & \textbf{Task Cat} & \textbf{Subtask} \\
\midrule
RemoteCount~\cite{liu2024remoteclip}  & O & DOTA & Multiple Cities & SC / FF & \textsf{X} & \checkmark & 1 & 1 \\
LEVIR-CC~\cite{liu2022remote} & O, BT & LEVIR-CD & 20 Regions in Texas & FF & \textsf{X} & \checkmark & 1 & 4 \\
DIOR-RSVG~\cite{DIOR-RSVG} & O & DIOR & $>$80 Countries & BBox & \textsf{X} &  \checkmark& 1 & 3 \\
RSVQA-LR/HR~\cite{lobry2020rsvqa} & O & Sentinel-2, USGS & Netherlands \& USA & FF & \checkmark &\textsf{X}  &5  & 5 \\
RSIEval~\cite{hu2025rsgpt} & O, PAN & DOTA & Multiple Cities & FF & \checkmark & \checkmark & 4 & 10 \\
LHRS-Bench~\cite{muhtar2024lhrs} & O & GE + OSM & N/A & SC / MCQ & \checkmark & \checkmark & 5 & 11 \\
EarthGPT~\cite{zhang2024earthgpt} & O, IR, SAR & DRSD & N/A & FF, BBox & \textsf{X} & -- & 5 &7 \\
SkyEyeGPT~\cite{zhan2025skyeyegpt} & O, V & DRSD & N/A & FF, MCQ & \textsf{X}& \checkmark & 6 & 6 \\
VRSBench~\cite{li2024vrsbench} & O & DOTA, DIOR & Multiple Regions & FF, BBox & \checkmark & \checkmark & 3 & 3 \\
EarthVQA~\cite{wang2024earthvqa} & O & LoveDA, WV-3 & 18 Regions in 3 Cities & FF & \checkmark & \textsf{X} & 6 & 6 \\
FineGrip~\cite{zhao2024panoptic}  & O & MAR20 & N/A & FF, Seg & \textsf{X} & \checkmark & 3 & 3 \\
GeoChat-Bench~\cite{kuckreja2024geochat} & O & SAMRS / DRSD & Multiple Regions & FF, BBox & \textsf{X} & \checkmark & 3 & 9 \\
UrBench~\cite{UrBench} & O & OSM/GE/DRSD & N/A & MCQ & \checkmark & \checkmark & 4 & 14 \\
SARLANG-1M~\cite{wei2026sarlang} & O, SAR & SpaceNet6/DFC2023/DRSD & 59 cities worldwide & FF (Caption, QA Pair) & \textsf{X} & \textsf{X} & 2 & 2 \\
LHRS-Align-Recap~\cite{LHRS-Bot-Nova} & O & OSM & N/A & FF (Caption) & \textsf{X} & \textsf{X} & 1 & 1 \\

DynamicVL~\cite{DynamicVL} & O & Diverse Platforms & 42
major cities in the U.S. & MCQ, FF & \checkmark & \checkmark & 6 & 6 \\

GEOBench-VLM~\cite{danish2025geobenchvlm} & O, MS, SAR, BT, MT & DRSD & Multiple Regions & MCQ, BBox, Seg & \checkmark & \checkmark & 8 & 31 \\
CHOICE~\cite{CHOICE} & O, MI, BT, MT & Diverse Platforms & 50 Cities Worldwide & MCQ, BBox, Seg & \checkmark & \checkmark & 6 & 23 \\
\midrule
\rowcolor[gray]{0.9} 
\textbf{OmniEarth (Ours)} & \textbf{O, MS, SAR, BT, MT} & \textbf{DRSD + JL-1 + Exclusive } & \textbf{7 Continents 400+ Cities} & \textbf{FF, MCQ, BBox, Seg} & \checkmark & \checkmark(all) & \textbf{8} & \textbf{28} \\
\bottomrule
\end{tabular}}
\flushleft{ O=Optical, PAN=Panchromatic, MS=Multi-spectral, IR=Infrared, SAR=Synthetic Aperture Radar, V=Video, MI=Multi-image, BT=Bi-Temporal, MT=Multi-temporal; DRSD=Diverse RS Datasets, OSM=OpenStreetMap, GE=Google Earth, WV-3=WorldView-3; MCQ=Multiple Choice, SC=Single Choice, FF=Free-Form, BBox=Bounding Box, Seg=Segmentation Mask.}
\label{tab:benchmarks}
\end{table*}

To bridge this gap, we propose OmniEarth, as illustrated in Figure~\ref{fig:examples}, an evaluation benchmark designed to systematically assess the capabilities of RSVLMs in Earth observation tasks. Specifically, we identify regions exhibiting significant long-term changes using World Imagery Wayback and conduct dedicated data acquisition to capture meaningful temporal dynamics. To ensure modality diversity, we collect SAR imagery from Capella and manually align it with corresponding optical footprints in Google Earth, forming high-fidelity SAR–RGB image pairs. To verify reasoning depth, OmniEarth includes tasks such as Geometric Measurement and Planning Suggestions, in which each image is queried using five semantically equivalent questions. This design enforces logical consistency across different linguistic formulations and encourages models to ground their predictions in visual evidence rather than relying on linguistic shortcuts~\cite{jin2023shortcutlens}, thereby preventing correct responses obtained without image access. To mitigate evaluation bias arising from data leakage, we manually curate high-resolution imagery from over 400 cities across seven continents and incorporate proprietary JL-1 satellite data, ensuring both data precision and the reliability of zero-shot evaluation.

Motivated by these challenges, we evaluate a range of state-of-the-art VLMs on OmniEarth, as shown in Table~\ref{tab:benchmarks}, including commercial foundation models, open-source general-purpose VLMs, and remote sensing–specific VLMs. Our experimental analysis leads to four key findings: \textbf{(1)} existing VLMs perform well on some image-level perception tasks but struggle with fine-grained perception that requires precise localization and segmentation; \textbf{(2)} reasoning capability remains limited, particularly for tasks involving temporal dependencies and domain-specific knowledge; \textbf{(3)} model robustness is weak under degraded image conditions and unseen modalities, with consistent performance degradation under noise, blur, and cross-modal RGB–SAR settings; \textbf{(4)} blind evaluations show that many RSVLM predictions rely more on textual than visual information, indicating weak visual grounding. These findings highlight both the progress achieved and the key gaps that remain in developing truly general and reliable RSVLMs. We summarize our contributions as follows:

\begin{itemize}
    \item We introduce \textbf{OmniEarth}, a benchmark for systematic evaluation of VLMs on Earth observation tasks, including \textbf{28} fine-grained tasks across perception, reasoning, and robustness, with bias-aware protocols on \textbf{9,275} quality-controlled images including proprietary JL-1 satellite imagery and \textbf{44,210} manually verified instructions.
    \item We evaluate \textbf{19} state-of-the-art VLMs across contrastive, general-purpose, and remote sensing–specific models, providing a comparative analysis of their geospatial capabilities and limitations.
    \item We evaluate VLMs on a broad range of geospatial tasks, covering multi-level perception, spatial and temporal reasoning, geographic application reasoning, and robustness under noisy and mismatched conditions, providing key insights for improving VLMs in geospatial applications.
\end{itemize}

\section{Related Works}
\subsection{Vision-Language Models in Remote Sensing} 
In recent years, RSVLMs~\cite{hu2025rsgpt,yao2025remotesam,zhu2025skysense,li2025segearth,luo2024skysensegpt,yao2025falcon,liu2024remoteclip} have developed rapidly, with an increasing number of representative models proposed. From a modeling perspective, existing RSVLMs can be broadly categorized into contrastive, conversational, and generative paradigms, among which conversational models that incorporate large language models for natural language interaction have received growing attention. Early works such as GeoChat~\citep{kuckreja2024geochat}, EarthGPT~\citep{zhang2024earthgpt} and LHRS-Bot~\cite{muhtar2024lhrs} demonstrated the feasibility of enabling geospatial dialogue and handling a variety of remote sensing queries~\cite{ye2025towards}. Subsequent models, including EarthDial~\citep{soni2025earthdial}, VHM~\citep{pang2025vhm}, SkyMoE~\citep{liu2025skymoe}, and GeoDiT~\citep{liu2025geodit}, further explored architectural designs, interaction mechanisms, and multi-task modeling strategies. In addition, RS-Thinker~\cite{rsthinker} introduced a perceptually grounded geospatial chain-of-thought framework that explicitly structures remote sensing analysis into task planning, visual evidence collection, and conclusion synthesis, providing an alternative approach for modeling reasoning processes in RSVLMs. The rapid development of RSVLMs highlights the need for objective, diverse, and systematic benchmarks to evaluate their capabilities.

\subsection{Benchmarks for RSVLMs}
Currently, relatively few benchmarks have been developed to evaluate remote sensing vision–language models for geospatial application tasks~\cite{dang2025benchmark}. As shown in Table~\ref{tab:benchmarks}, benchmarks such as EarthVQA~\cite{wang2024earthvqa}, LHRS~\cite{muhtar2024lhrs}, EarthDial~\cite{soni2025earthdial}, and GeoChat~\cite{kuckreja2024geochat} are often released alongside their models, a practice that may introduce potential data leakage risks. VRSBench~\cite{li2024vrsbench} designs evaluation tasks including caption, grounding, and VQA; however, its evaluation dimensions and problem scale remain relatively limited and may not keep pace with the rapid development of RSVLMs. CHOICE~\cite{CHOICE} constructs multiple-choice evaluation tasks with definitive answers across 23 tasks, but provides limited support for long-term temporal analysis and pixel-level fine-grained interpretation. To assess large-scale RSI understanding, benchmarks such as RSHR-Bench~\cite{dang2025benchmark}, XLRS-Bench~\cite{wang2025xlrs}, and LRS-VQA~\cite{luo2025large} adopt semi-automated construction strategies on wide-area data. In addition, GeoBench-VLM~\cite{danish2025geobenchvlm} focuses on specific geospatial tasks, with evaluations largely concentrated on fine-grained classification settings, such as multiple subcategories for object localization and counting, resulting in relatively limited task diversity at a higher level. To address image quality issues, REOBench~\cite{REOBench} introduces robustness benchmarks under various image degradation conditions. Therefore, there is a clear need for a unified and systematic evaluation framework that integrates multi-sensor data, multi-resolution imagery, a broad and fine-grained set of geospatial tasks, as well as deep reasoning and robustness analysis within a single, practically grounded benchmark.

\section{OmniEarth}

\begin{figure*}[htbp]
    \centering
    \includegraphics[width=\linewidth]{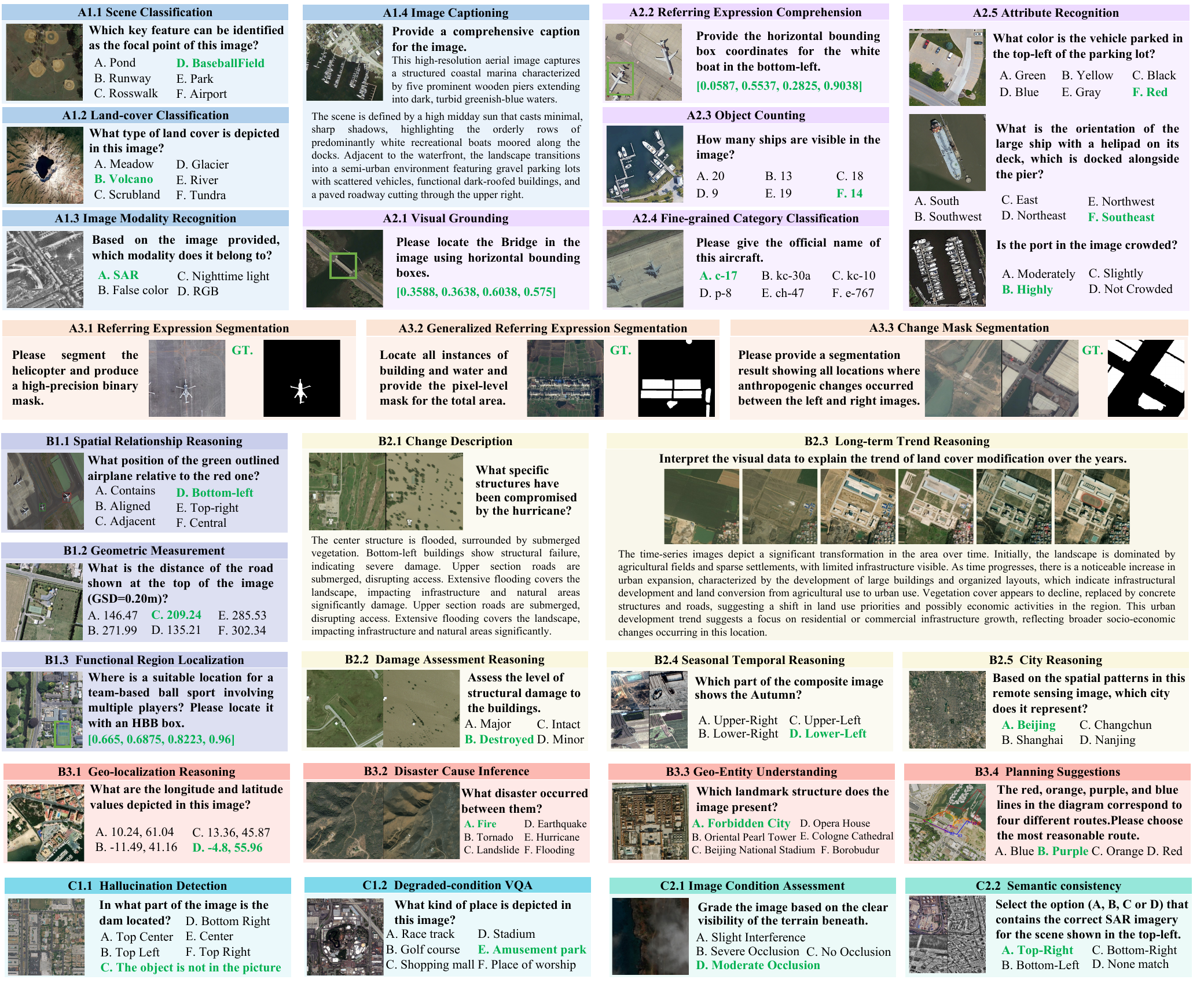}
    \vspace{-0.35cm}
    \caption{OmniEarth comprehensively covers 28 fine-grained tasks categorized into 3 categories: A. Perception, B. Reasoning, and C. Robustness.}
    \label{fig:examples}
\end{figure*}

\begin{figure*}[htbp]
  \centering
  \begin{minipage}{0.6\linewidth}
    \centering
    \includegraphics[width=\linewidth]{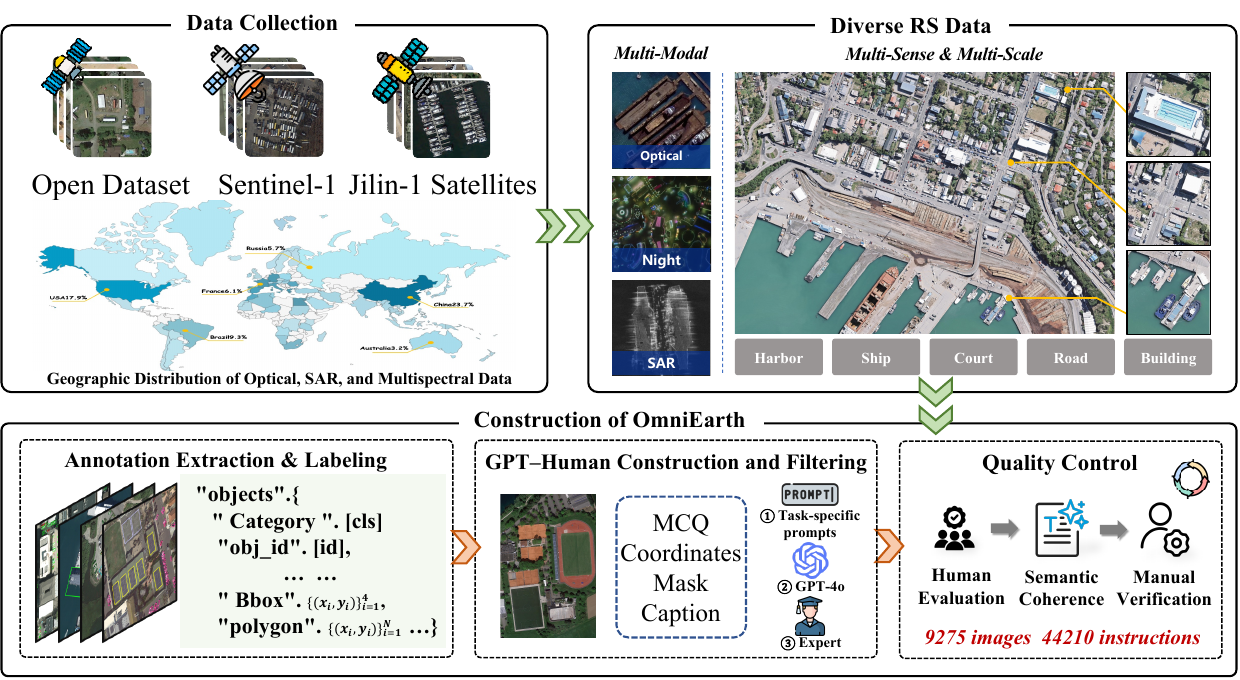}
    \vspace{-0.35cm}
    \caption{Overview of the construction of OmniEarth}
    \label{fig:pipline}
  \end{minipage}\hfill
  \begin{minipage}{0.37\linewidth}
    \centering
    \includegraphics[width=\linewidth]{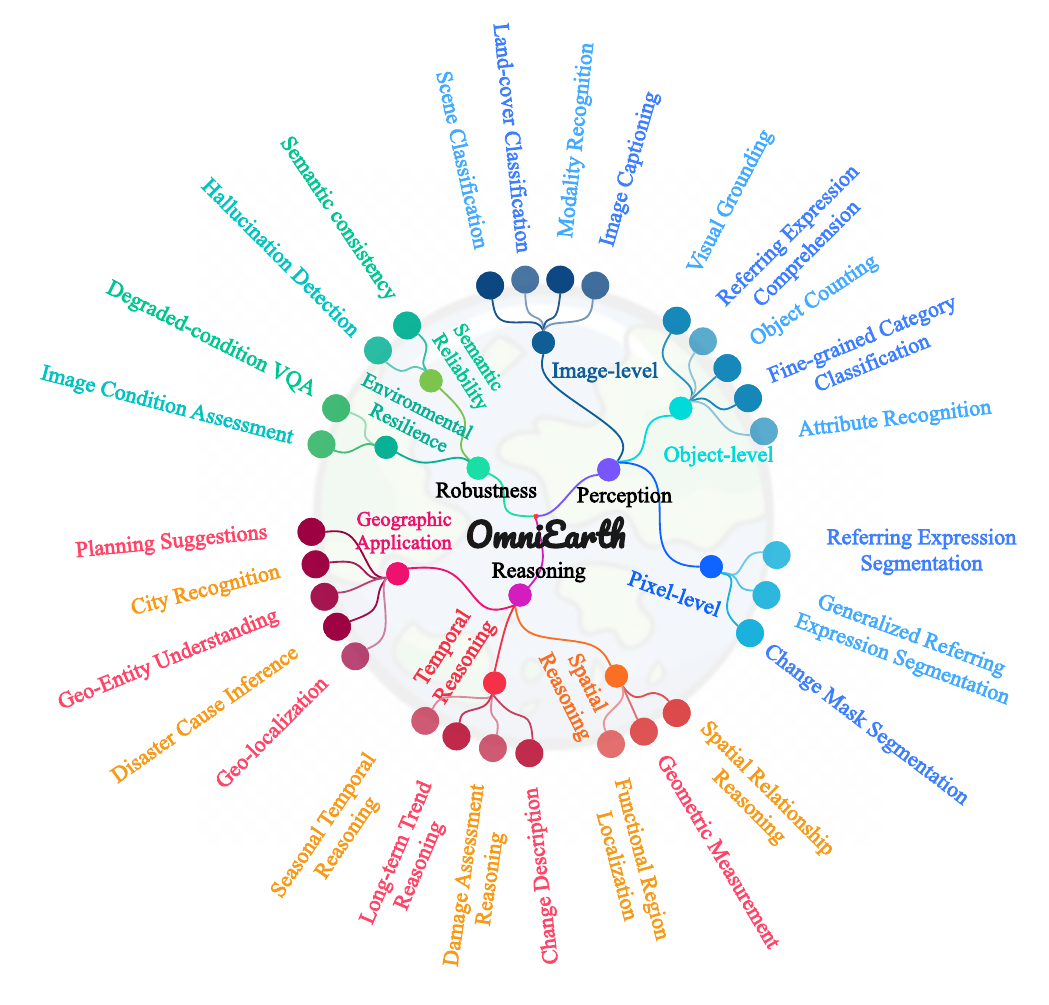}\\
    \vspace{-0.35cm}
    \caption{Hierarchy Taxonomy}
    \label{fig:tax}
  \end{minipage}
\end{figure*}

As shown in Figure~\ref{fig:tax}, OmniEarth adopts a hierarchical taxonomy to structure the evaluation of RSVLMs, organizing 28 fine-grained tasks into three dimensions: Perception, Reasoning, and Robustness. At the perception level, tasks are categorized into image-level, instance-level, and pixel-level evaluation. At the reasoning level, tasks are further divided into spatial reasoning, temporal reasoning, and geospatial reasoning. Robustness is assessed from two aspects: environmental resilience and semantic reliability. Figure~\ref{fig:examples} illustrates representative examples for all 28 fine-grained tasks in OmniEarth. More detailed task descriptions are given in Appendix~\ref{suppl:tasks}.

\vspace{-3pt}
\subsection{Perception}
Perception constitutes the foundational level of the OmniEarth hierarchy and focuses on evaluating a model’s ability to extract visual and structural information from remote sensing imagery (RSI). Given the diversity of satellite sensors and the substantial domain gap between natural images and remote sensing observations, this dimension emphasizes perception beyond standard computer vision settings. We organize perceptual capability into three granularities: \textbf{image-level}, \textbf{instance-level}, and \textbf{pixel-level} perception, covering global scene understanding, object-centric recognition, and dense spatial interpretation.

\textbf{Image-level perception} evaluates overall understanding of geospatial scenes and includes four tasks. Scene Classification (SC) assesses recognition of artificial objects such as aircraft, vehicles, bridges, and roads, while Land-cover Classification (LCC) focuses on natural surface types including forests, glaciers, bare land, and volcanic areas. Image Modality Recognition (IMR) examines whether models can distinguish sensing modalities, including optical, SAR, multispectral, and nighttime imagery. Image Captioning (IC) requires generating natural language descriptions that summarize the overall scene context.

\textbf{Instance-level perception} focuses on object-centric understanding and localization under open-vocabulary settings. Visual Grounding (VG) requires localizing a target object in scenes with a single salient instance, whereas Referring Expression Comprehension (REC) extends this to multi-object scenarios using natural language descriptions. Object Counting (OC) evaluates numerical accuracy for both general objects and disaster-related structures. Fine-grained Category Classification (FCC) tests taxonomic specificity, such as distinguishing aircraft models, while Attribute Recognition (AR) examines the recognition of object properties including orientation, size, and occupancy status.

\textbf{Pixel-level perception} evaluates dense spatial understanding and boundary precision. Referring Expression Segmentation (RES) requires producing pixel-accurate masks for linguistically specified objects, and Generalized Referring Expression Segmentation (GRES) extends this to multiple or absent targets. Change Mask Segmentation (CMS) addresses temporal perception by requiring pixel-level delineation of changes from bi-temporal imagery.

\vspace{-3pt}
\subsection{Reasoning}

Reasoning evaluates whether a model can answer geospatial questions by leveraging RSI together with domain-specific knowledge. Unlike perception tasks that focus on visual recognition, this dimension emphasizes the ability of RSVLMs to perform relational analysis, temporal interpretation, and application-oriented inference commonly required in RS interpretation. Accordingly, reasoning tasks are grouped into three categories: \textbf{spatial reasoning}, \textbf{temporal reasoning}, and \textbf{geographic application reasoning}.

\textbf{Spatial reasoning} evaluates a model’s understanding of relative positions, geometric relationships, and functional layouts. Spatial Relationship Reasoning (SRR) requires identifying positional relations such as adjacency, containment, and surrounding patterns among objects. Geometric Measurement (GM) evaluates quantitative inference by requiring the estimation of physical properties, including building areas, road lengths, and distances between objects. Functional Region Localization (FRL) extends spatial reasoning to implicit functionality, where models must identify regions suitable for specific activities, such as emergency shelters or irrigated farmland, based on visual cues.

\textbf{Temporal reasoning} focuses on interpreting changes across multiple times. Change Description (CD) requires models to identify and describe differences between bi-temporal images in natural language. Damage Assessment Reasoning (DAR) evaluates the interpretation of disaster impacts, including infrastructure damage and road accessibility. Long-term Trend Reasoning (LTR) addresses multi-temporal analysis by requiring models to infer broader trends, such as urban expansion or vegetation evolution, from image sequences. Seasonal Temporal Reasoning (STR) evaluates seasonal recognition by requiring models to identify spring, summer, autumn, and winter from a four-quadrant composite image.

\textbf{Geographic application reasoning} targets high-level tasks that combine visual interpretation with domain-specific geographic knowledge. Geo-localization (GL) requires identifying climatic zones or approximate locations based on scene characteristics. Disaster Cause Inference (DCI) evaluates the ability to infer likely causes of events such as floods, fires, or industrial accidents. Geo-Entity Understanding (GEU) and City Recognition (CR) assess the recognition of prominent geographic entities and major cities. Planning Suggestions (PS) evaluates whether models can select or recommend feasible routes under given constraints.

\subsection{Robustness}
Robustness evaluates whether model predictions remain consistent under degraded input conditions and misleading task settings. RSI is frequently affected by sensor characteristics, atmospheric conditions, and acquisition constraints, making performance under low-quality inputs an important aspect of evaluation. Accordingly, this dimension focuses on model behavior under signal degradation and semantic ambiguity, and the tasks are grouped into two categories: \textbf{environmental resilience} and \textbf{semantic reliability}.

\textbf{Environmental resilience} focuses on model performance under adverse imaging conditions and signal degradation. Image Condition Assessment (ICA) requires models to judge image usability and identify environmental factors such as cloud cover, haze, snow, and dense urban occlusion. Degraded-condition VQA (DVQA) evaluates whether models can answer questions consistently when input images are affected by physical or digital degradations. These include geometric distortions and signal-level noise such as Gaussian blur, salt-and-pepper noise, compression artifacts, and atmospheric haze. By applying multiple degradation types across different acquisition conditions, this category measures the sensitivity of model predictions to realistic sensing noise.

\textbf{Semantic reliability} evaluates whether model predictions remain logically consistent under misleading or adversarial semantic settings. Hallucination Detection (HD) evaluates a model’s ability to reject incorrect assumptions by requiring it to identify non-existent objects, verify factual scene descriptions, and recognize missing elements through comparative reasoning. Semantic Consistency (SEC) further assesses cross-modal alignment by presenting a single optical image together with multiple candidate SAR tiles, including cases without a valid match. Models are required to identify the correct SAR–optical correspondence based on structural consistency rather than visual similarity alone. 

\section{Construction of OmniEarth}

OmniEarth includes 28 fine-grained tasks evaluated under four output formats. Figure~\ref{fig:pipline} illustrates the overall construction of OmniEarth. First, for most tasks, a multiple-choice question (MCQ) format is used, where each problem instance is defined as $P_i = [Q_i, C_i, I_i, G_i]$. Here, $Q_i$ denotes the textual query, which typically consists of five semantically consistent questions for evaluating output consistency, $C_i$ is a set of $n$ candidate choices (typically $n \in [2,6]$), $I_i$ is the associated RSI, and $G_i$ is the ground-truth label. Notably, for hallucination detection tasks, a multi-answer MCQ format with a variable number of choices is used, and scoring is defined as follows: a prediction receives a score of 1 if all correct options are selected and no incorrect option is included, 0.5 if a subset of correct options is selected without any incorrect option, and 0 otherwise. This design is motivated by an empirical observation that, under text-only input settings, some models can infer the correct answer by eliminating options using semantic correlations rather than image. Second, for localization tasks, the choice set $C_i$ is omitted and $G_i$ is defined as the ground-truth bounding box coordinates. Third, for segmentation tasks, models output a binary mask aligned with the input image resolution, with $G_i$ defined as the pixel-wise annotation. Finally, for open-ended descriptive tasks, models generate natural language outputs without predefined candidate choices.

\subsection{Data Source}

From a global perspective, RSI vary across different regions and land-cover categories. Existing evaluation benchmarks often concentrate on a limited set of areas, such as ports, cities, and airports, and in some cases, multiple tasks rely on data from the same geographic region. To address this issue, OmniEarth includes images from all continents except Antarctica and covers more than 400 cities worldwide. For tasks that involve long-term changes or large-scale spatial context, such as long-term trend inference and geographic entity understanding, additional images are sampled from random regions worldwide to increase scene coverage. OmniEarth contains 9,275 images collected from two sources. The first source consists of open-source data~\cite{christie2018functional,xia2017aid,liu2024cross,LI2020296,xia2018dota,wenqi2024mar20,gupta2019xbd,essd-17-6217-2025} that are cleaned by removing all original labels and task definitions, while retaining only raw pixel data and basic geometric annotations. The second source includes imagery from the JL-1 constellation, Sentinel-1, Sentinel-2, and Google Earth Engine. \textit{The JL-1 images are publicly released for the first time and are less likely to appear in existing model training data.} Image resolution is set based on task requirements, with ground sampling distances ranging from 0.05 to 30 meters. More detailed data sources are provided in Appendix~\ref{suppl:data}. 

\subsection{Benchmark Pipline}
OmniEarth adopts two complementary approaches for selecting task instances and generating annotations. The first approach relies on existing remote sensing datasets, including classification, detection, and segmentation datasets, from which original annotations are extracted. Although these annotations are generally accurate, all selected samples are manually inspected to remove ambiguous or low-quality cases. The second approach is task-driven and involves collecting task-specific data followed by manual annotation. These two approaches are used together to support tasks with different data availability and annotation requirements.

\textbf{Dataset-driven Construction.} For tasks with reliable open-source annotations, such as scene classification, attribute recognition, and basic spatial reasoning, images and corresponding ground-truth labels are collected from existing datasets~\cite{christie2018functional,xia2017aid,sun2022fair1m,9044734,8444434,9099032,Chen2020}. All samples are manually reviewed to remove ambiguous or low-quality cases. Based on the verified data, task-specific prompts are designed, and models such as GPT-4o~\cite{hurst2024gpt4o} are used to generate question descriptions and candidate options. Since these models are mainly used for text generation rather than precise numerical or spatial annotation~\cite{10678128}, original annotations are directly adopted as answers for tasks requiring high accuracy, including object counting, localization, and segmentation. In these cases, LLMs~\cite{hurst2024gpt4o} are only used to generate question text and distractor options. For scene classification tasks, the original category labels serve as correct answers, and distractors are selected from labels of other samples within the same task to ensure visual distinguishability and reasonable option design.

\textbf{Task-driven Exclusive Construction.} For tasks lacking high-quality public data, including fine-grained scene classification, seasonal reasoning, geo-entity understanding, and city recognition, we collect exclusive imagery from the JL-1 satellite and retrieve time-stamped data from World Imagery Wayback~\cite{googleearth,livingatlas,capellaspace}. Image locations are manually verified using geographic references such as Google Maps, and one-to-one labeling is performed based on building coordinates or known geographic entities. Overall, We perform multiple prompt iterations with an overall rejection rate exceeding 60\%. To ensure evaluation robustness, distractors are specifically designed to deviate from the correct dimensions by more than 30\%. Implementation details and corresponding prompt designs for each task are provided in the appendix.

\subsection{Human Verification}

Quality control is maintained through a three-group, three-round cross-validation protocol. Each group consists of two professionals who must reach a unanimous decision for a sample to be accepted. Consistency is verified by measuring response variance across multiple rounds. The near-zero variance confirms that the five questions are semantically equivalent and effectively isolate model performance from prompt influence, providing a deterministic measure of geospatial grounding.

\begin{table*}[!t]
    \caption{Comprehensive evaluation of 19 vision-language models across the OmniEarth hierarchy.}
    \vspace{-0.35cm}
    \centering
    \label{tab:FullResults_Final_V5}
    \newcommand{\first}[1]{\cellcolor{red!20!}\textbf{#1}}
    \newcommand{\second}[1]{\cellcolor{blue!15!}#1}
    
    \resizebox{\textwidth}{!}{
        \begin{tabular}{l | cccccccccccc | cccccccccccc | cccc}
            \toprule
            \multirow{2}{*}{\textbf{Method}} & \multicolumn{12}{c|}{\textbf{Perception (12 Tasks)}} & \multicolumn{12}{c|}{\textbf{Reasoning (12 Tasks)}} & \multicolumn{4}{c}{\textbf{Robustness (4 Tasks)}} \\
            \cmidrule(lr){2-13} \cmidrule(lr){14-25} \cmidrule(lr){26-29}
            & \textbf{\rotatebox{60}{SC}} & \textbf{\rotatebox{60}{LCC}} & \textbf{\rotatebox{60}{IMR}} & \textbf{\rotatebox{60}{IC}} & \textbf{\rotatebox{60}{VG}} & \textbf{\rotatebox{60}{REC}} & \textbf{\rotatebox{60}{OC}} & \textbf{\rotatebox{60}{FCC}} & \textbf{\rotatebox{60}{AR}} & \textbf{\rotatebox{60}{RES}}& \textbf{\rotatebox{60}{GRES}} & \textbf{\rotatebox{60}{CMS}} & \textbf{\rotatebox{60}{SRR}} & \textbf{\rotatebox{60}{GM}} & \textbf{\rotatebox{60}{FRL}} & \textbf{\rotatebox{60}{CD}} & \textbf{\rotatebox{60}{DAR}} & \textbf{\rotatebox{60}{LTR}} & \textbf{\rotatebox{60}{STR}} & \textbf{\rotatebox{60}{GL}} & \textbf{\rotatebox{60}{DCI}} & \textbf{\rotatebox{60}{GEU}} & \textbf{\rotatebox{60}{CR}} & \textbf{\rotatebox{60}{PS}} & \textbf{\rotatebox{60}{ICA}} & \textbf{\rotatebox{60}{DVQA}} & \textbf{\rotatebox{60}{HD}} & \textbf{\rotatebox{60}{SEC}} \\
            \midrule

            \rowcolor{gray!10}\multicolumn{29}{l}{\textit{Specialized Encoders}} \\
            SkyCLIP-ViT-B~\cite{wang2024skyscript}  & 8.4 & 24.1 & 25.0 & - & - & - & 10.0 & 8.0 & 21.3 & - & - & - & 25.4 & 18.7 & - & - & 30.4 & - & 20.8 & 26.3 & 24.2 & 13.3 & 17.4 & 26.5 & 30.5 & 15.4 & - & 20.0 \\
            RemoteCLIP-ViT-B~\cite{liu2024remoteclip}  & 49.8 & 84.8 & 45.7 & - & - & - & 49.9 & 26.7 & 27.2 & - & - & - & 27.5 & 14.1 & - & - & 32.0 & - & 20.8 & 30.6 & 22.6 & 51.0 & 25.6 & 23.2 & 28.4 & 47.6 & - & 25.8 \\
            GeoRSCLIP~\cite{zhang2024rs5m}  & \first{72.6} & 81.1 & 54.3 & - & - & - & 28.6 & 40.1 & 39.2 & - & - & - & 7.0 & 10.5 & - & - & 32.1 & - & 19.8 & 29.4 & 41.3 & 84.0 & 35.7 & 24.6 & 24.0 & \first{57.8} & - & 29.6 \\
            \midrule

            \rowcolor{gray!10}\multicolumn{29}{l}{\textit{General Close-source Models}} \\
            GLM-4.6V~\cite{glm}  & 60.1 & 70.9 & 67.3 & 120.4 & 60.2 & \first{63.0} & 26.3 & 39.0 & 38.8 & 5.9 & 7.3 & \second{10.9} & 81.0 & 16.2 & 62.2 & 85.0 & 39.6 & 82.7 & 38.1 & \second{54.3} & \second{43.8} & 79.7 & 61.3 & \second{39.8} & 37.2 & 35.9 & 15.2 & 18.1 \\
            Claude-sonnet-4~\cite{anthropic2025claude4}  & 62.7 & 81.3 & 82.9 & 137.3 & 14.1 & 2.0 & 50.1 & 49.4 & \second{58.8} & 3.3 & 4.2 & 9.9 & 85.2 & 15.8 & 28.5 & 96.8 & 35.5 & 60.7 & 28.7 & 34.3 & 26.9 & 84.4 & 38.7 & 32.1 & \second{63.3} & 42.2 & 62.0 & 57.2 \\
            Gemini-2.0-Flash~\cite{Gemini20Flash}  & \second{71.3} & 82.8 & \second{85.5} & \second{150.7} & 30.1 & 11.8 & 48.2 & \first{63.4} & \first{58.9} & 10.2 & 10.8 & 1.9 & 81.3 & 12.6 & 19.2 & 95.1 & \second{39.9} & 87.9 & 26.1 & 43.3 & 40.6 & \first{96.0} & \first{78.6} & \first{45.6} & 56.0 & \second{48.2} & 61.4 & \second{63.8} \\
            GPT-4o~\cite{hurst2024gpt4o}  & 65.8 & \first{89.3} & \first{87.1} & \first{151.9} & 16.6 & 3.6 & 33.8 & 51.6 & 55.6 & 4.3 & 4.9 & 10.7 & 74.6 & \second{33.7} & 13.4 & \first{113.3} & 31.9 & 68.9 & 17.8 & 25.7 & 41.2 & \second{91.8} & 53.8 & 24.2 & 60.0 & 46.4 & 52.0 & \first{73.2} \\
            \midrule

            \rowcolor{gray!10}\multicolumn{29}{l}{\textit{General Open-source Models}} \\
            Qwen2.5-VL-72B~\cite{Qwen2.5-VL}  & 59.8 & 80.5 & 75.5 & 80.0 & 52.9 & \second{61.0} & 48.7 & \second{54.6} & 55.3 & 3.8 & 2.6 & 6.7 & \second{87.4} & 28.1 & 44.8 & \second{105.1} & \first{40.6} & 100.7 & \second{38.9} & \first{62.7} & \first{51.0} & 83.7 & \second{73.8} & 24.2 & \first{67.8} & 42.5 & \first{75.8} & 46.3 \\
            Qwen3-VL-8B~\cite{yang2025qwen3}  & 58.7 & 80.6 & 73.4 & 137.8 & \second{64.8} & 56.1 & 47.7 & 52.5 & 53.0 & 7.2 & 4.1 & 0.5 & 79.6 & 24.1 & \second{62.8} & 95.1 & 32.6 & 91.9 & 33.9 & 54.1 & 31.3 & 78.7 & 68.1 & 33.6 & 45.8 & 42.9 & 69.4 & 39.0 \\
            Qwen3-VL-32B~\cite{yang2025qwen3}  & 60.2 & 82.2 & 76.4 & 131.8 & 56.1 & 41.0 & 57.6 & 50.1 & 55.8 & 6.6 & 4.5 & 6.4 & \first{88.6} & 17.4 & \first{69.8} & 92.2 & 38.7 & 86.7 & \first{41.1} & 51.2 & 35.9 & 75.7 & 66.1 & 31.8 & 52.4 & 45.1 & \second{73.0} & 39.8 \\
            InternVL3-8B~\cite{zhu2025internvl3}  & 63.2 & 74.0 & 65.0 & 127.6 & 19.6 & 3.6 & 48.2 & 41.0 & 51.3 & 3.8 & 2.4 & 7.2 & 80.1 & \first{37.0} & 2.9 & 101.0 & 37.7 & 109.2 & 30.0 & 46.6 & 33.5 & 74.8 & 42.6 & 31.5 & 53.4 & 40.0 & 65.2 & 39.1 \\
            InternVL3.5-8B~\cite{wang2025internvl3_5}  & 51.2 & 78.2 & 64.8 & 129.4 & 53.1 & 18.9 & \first{60.5} & 36.6 & 50.8 & 14.1 & 12.5 & 9.1 & 86.0 & 29.3 & 29.1 & 85.7 & 25.6 & \second{110.8} & 25.2 & 41.3 & 27.2 & 56.8 & 35.9 & 30.3 & 50.9 & 42.1 & 63.1 & 40.4 \\
            InternVL3.5-38B~\cite{wang2025internvl3_5}  & 66.4 & \second{85.0} & 72.7 & 132.9 & \first{69.8} & 13.8 & \second{60.4} & 48.3 & 53.8 & \second{17.0} & \second{13.5} & \first{12.8} & 86.2 & 23.2 & 38.4 & 89.7 & 32.6 & \first{111.1} & 33.2 & 51.2 & 41.7 & 74.0 & 41.9 & 31.1 & 62.3 & 46.6 & 69.2 & 48.8 \\
            \midrule

            \rowcolor{gray!10}\multicolumn{29}{l}{\textit{Remote Sensing Specialized Models}} \\
            VHM-7B~\cite{pang2025vhm}  & 36.4 & 56.5 & 29.5 & 113.0 & 41.6 & 12.9 & 18.8 & 19.7 & 23.6 & 13.3 & 10.8 & 1.2 & 15.6 & 14.1 & 2.9 & 93.8 & 28.6 & 100.6 & 27.2 & 24.7 & 16.0 & 43.3 & 15.9 & 24.3 & 27.1 & 37.8 & 26.4 & 26.1 \\
            GeoChat~\cite{kuckreja2024geochat}  & 43.9 & 72.9 & 29.6 & 138.1 & 4.3 & 1.1 & 17.8 & 23.6 & 24.5 & 6.8 & 9.5 & 4.3 & 14.9 & 5.7 & 2.9 & 102.4 & 29.7 & 99.0 & 26.5 & 25.8 & 20.3 & 48.0 & 16.2 & 25.4 & 25.8 & 25.3 & 31.0 & 25.9 \\
            EarthDial-RGB~\cite{soni2025earthdial}  & 58.6 & 78.9 & 39.6 & 72.7 & 54.7 & 40.5 & 37.2 & 23.8 & 40.2 & \first{23.0} & \first{17.6} & 3.9 & 14.5 & 12.7 & 18.0 & 68.1 & 30.0 & 74.6 & 26.2 & 31.2 & 12.0 & 39.5 & 22.0 & 28.5 & 35.9 & 39.6 & 37.8 & 20.0 \\
            EarthDial-MS~\cite{soni2025earthdial}  & 41.4 & 46.9 & 26.8 & 66.3 & 0.0 & 0.0 & 19.0 & 21.2 & 24.3 & 0.0 & 0.0 & 0.0 & 14.9 & 8.7 & 0.0 & 71.5 & 30.7 & 59.5 & 22.8 & 22.6 & 14.8 & 21.8 & 16.3 & 29.8 & 19.1 & 24.2 & 25.6 & 21.4 \\
            GeoLLaVA-8K~\cite{wang2025geollava}  & 0.0 & 18.2 & 27.7 & 0.0 & 0.0 & 0.0 & 19.5 & 18.0 & 23.2 & 0.0 & 0.0 & 0.0 & 35.7 & 16.7 & 0.0 & 0.0 & 25.8 & 0.0 & 28.7 & 24.5 & 21.0 & 17.2 & 22.2 & 25.3 & 24.8 & 23.6 & 19.9 & 27.6 \\
            SkySenseGPT~\cite{luo2024skysensegpt}  & 51.2 & 66.9 & 32.3 & 125.6 & 11.3 & 6.7 & 17.6 & 25.1 & 22.8 & 0.0 & 0.0 & 0.0 & 15.0 & 5.6 & 1.7 & 95.6 & 29.8 & 99.0 & 24.0 & 36.2 & 18.3 & 48.8 & 16.4 & 25.4 & 24.9 & 28.4 & 34.8 & 25.8 \\
            \bottomrule
        \end{tabular}
    }
    \vspace{2pt}
    \begin{flushleft}
    \footnotesize
    \textbf{Notes:} Task mappings are defined as: \textbf{Perception}: SC (Scene Classification), LCC (Land-cover Classification), IMR (Image Modality Recognition), IC (Image Captioning), VG (Visual Grounding), REC (Referring Expression Comprehension), OC (Object Counting), FCC (Fine-grained Category Classification), AR (Attribute Recognition), RES (Referring Expression Segmentation), GRES (Generalized Referring Expression Segmentation), CMS (Change Mask Segmentation); \textbf{Reasoning}: SRR (Spatial Relationship Reasoning), GM (Geometric Measurement), FRL (Functional Region Localization), CD (Change Description), DAR (Damage Assessment Reasoning), LTR (Long-term Trend Reasoning), Seasonal Temporal Reasoning (STR), GL (Geo-localization), DCI (Disaster Cause Inference), GEU (Geo-Entity Understanding), CR (City Recognition), PS (Planning Suggestions); \textbf{Robustness}: ICA (Image Condition Assessment), DVQA (Degraded-condition VQA), HD (Hallucination Detection), SEC (Semantic Consistency). Metric standardization: \textbf{acc@0.5} (IoU threshold 0.5) for VG, RES, and FRL; \textbf{CIDEr} for IC, CD, and LTR; \textbf{mIoU} for RES, GRES, and CMS;  other tasks use \textbf{Top-1 Accuracy} ($acc_{avg}$). CIDEr scores are scaled by 1000, all other values are scaled by 100. \textcolor{red}{Red} denotes the best performance, and \textcolor{blue}{Blue} denotes the second-best performance.

    \end{flushleft}
\end{table*}

\subsection{Evaluation Metrics}
We adopt task-specific metrics to evaluate model performance. Accuracy is reported for MCQ tasks. For localization tasks, detection performance is measured using precision at IoU@0.5. For segmentation tasks, mean Intersection-over-Union (mIoU) is reported. For image captioning tasks, caption quality is evaluated using CIDEr, BLEU-4, ROUGE-L, METEOR, and BERTScore.

\section{Evaluation Results}

\subsection{Experimental Setup}
We evaluate OmniEarth using 19 representative vision–language models, which are grouped into four categories according to their model type. First, to provide a reference for remote sensing visual encoding, we evaluate three RS-specific contrastive models: SkyCLIP~\cite{wang2024skyscript}, RemoteCLIP~\cite{liu2024remoteclip}, and GeoRSCLIP~\cite{zhang2024rs5m}. Second, we include four proprietary large-scale VLMs: GPT-4o~\cite{hurst2024gpt4o}, Gemini-2.0-Flash~\cite{Gemini20Flash}, Claude-3.5-Sonnet~\cite{Claude35Sonnet}, and GLM-4.6V~\cite{glm}. Third, we evaluate six open-source general-purpose VLMs, including the InternVL3~\cite{zhu2025internvl3} family (3B, 3.5B, and 38B variants), Qwen3-VL~\cite{yang2025qwen3} (8B and 32B), and Qwen2.5-VL-72B~\cite{Qwen2.5-VL}. Finally, we evaluate six remote sensing–specific VLMs that incorporate domain-adapted training or fine-tuning, including GeoChat~\cite{kuckreja2024geochat}, SkySenseGPT~\cite{luo2024skysensegpt}, VHM-7B~\cite{pang2025vhm}, GeoLLaVA-8K~\cite{wang2025geollava}, and models from the EarthDial~\cite{soni2025earthdial} series. All open-source models are evaluated on a server equipped with 8 NVIDIA H200 GPUs. All models are evaluated in a zero-shot setting using their default inference configurations, without task-specific tuning or prompt optimization, to ensure consistency and reproducibility. Detailed results are presented in Table~\ref{tab:FullResults_Final_V5}. Supplemental experiments regarding blind evaluations, visual gain analysis, and consistency measurements are provided in Appendix~\ref{suppl:exp}.

\subsection{Main results}

\textbf{Perception.} At the image level, general-domain VLMs such as GPT-4o, Gemini-2.0-Flash, and InternVL3.5-38B achieve strong performance on SC, LCC, and IMR, with accuracies ranging from around 65\% to nearly 90\%. In contrast, RSVLMs show weaker image-level perception, with SC and IMR accuracies mostly concentrated in the 25\%–45\% range. As evaluation shifts to finer-grained perception, performance degrades across all model families. At the object level, even the strongest models reach only moderate performance, with OC and AR scores generally around 50\%–60\%. The degradation becomes more pronounced at the pixel level: for segmentation tasks including RES, GRES, and CMS, most models achieve scores below 15\%, and many RSVLMs report near-zero results. Overall, these results indicate that fine-grained spatial understanding, particularly at object and pixel scales, remains a major challenge for current VLMs.

\textbf{Reasoning.} At the spatial reasoning level, general-domain VLMs perform well on qualitative tasks such as SRR, with Qwen3-VL-32B, InternVL3.5-38B, and Qwen2.5-VL-72B achieving over 85\%. However, performance drops markedly on quantitative tasks such as GM and FRL, where most models remain below 40\%, indicating limited geometric and metric reasoning ability. RSVLMs show consistently weaker spatial reasoning, with SRR and GM scores mostly below 20\%. At the temporal reasoning level, general-domain models and RSVLMs exhibit similar performance on descriptive tasks such as CD and LTR. However, the overall scores on these tasks remain moderate, indicating that the ability to finely describe temporal changes is still limited. For STR, all models perform poorly, with most results falling below 40. These findings suggest that neither general-domain models nor RSVLMs show strong fine-grained temporal reasoning capabilities; instead, current models rely more on coarse temporal pattern analysis rather than explicit and precise reasoning over temporal changes. Finally, in geographic application reasoning, general-domain models perform well on regional cognition tasks such as GEU and CR. In contrast, performance drops markedly on decision- and planning-oriented tasks such as GL and PS, with most models concentrated in the 20\%–35\% range. However, tasks that require deeper integration of geographic knowledge still show limited performance. Overall, these results suggest that effectively integrating spatial, temporal, and geographic cues for complex decision-making remains a significant challenge.

\textbf{Robustness.} In environmental resilience tests, general-domain VLMs demonstrate stronger robustness to image degradation. For instance, InternVL3.5-38B achieves 62.3\% accuracy on ICA, while RSVLMs such as GeoChat and VHM-7B remain around 25\%–27\%, indicating limited robustness under degraded visual conditions. A similar gap is observed on DVQA, where most general models score above 40\%, whereas RSVLMs remain below 30\%. In semantic reliability tasks, general VLMs again outperform RSVLMs. General models achieve substantially higher scores on HD. For example, Qwen2.5-VL-72B reaches 75.8\% and Gemini-2.0-Flash achieves 61.4\%, while RSVLMs remain below 40\%. For SEC, general models commonly exceed 40\%–70\%, whereas most RSVLMs cluster around 20\%–30\%, with some models exhibiting near-zero performance. Overall, these results indicate that current RSVLMs show weaker robustness and limited semantic reliability under noisy or inconsistent conditions.

\begin{figure}[t]
  \centering
  \begin{minipage}{\linewidth}
    \centering
    \includegraphics[width=\linewidth]{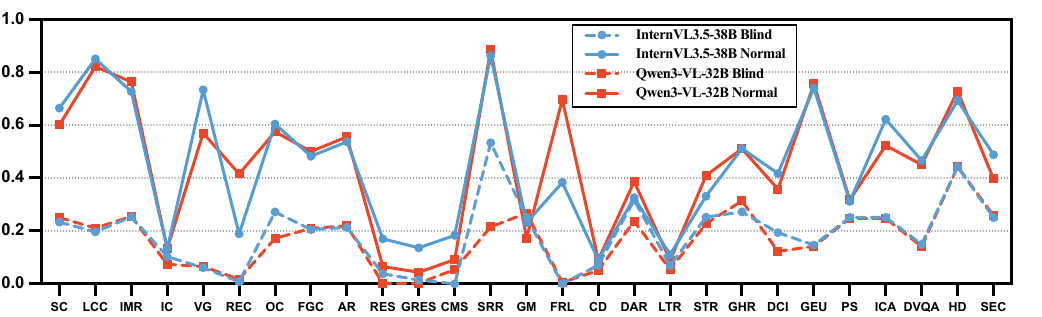}\\
    {\small (a) General Open Source Models}
  \end{minipage}
  \vspace{0.3em}
  \begin{minipage}{\linewidth}
    \centering
    \includegraphics[width=\linewidth]{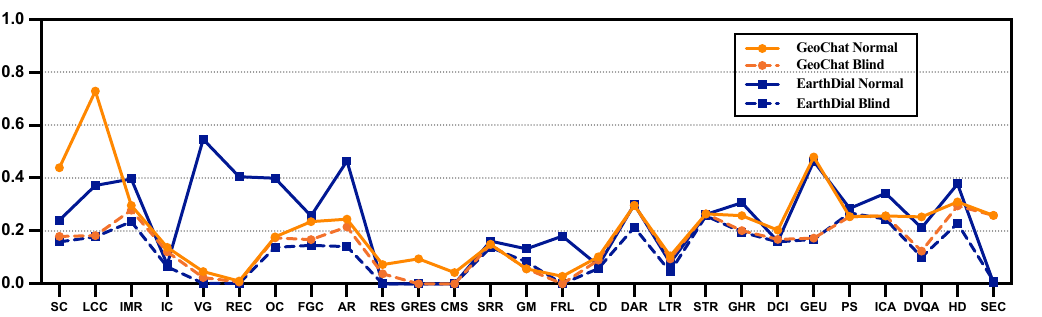}\\
    {\small (d) RS domain-specific Models}
  \end{minipage}
    \vspace{-0.35cm}
  \caption{Performance gaps between normal evaluation and blind evaluation for general open-source and RS domain-specific models.}
  \label{fig:blindtest}
\end{figure}

\subsection{Performance Analysis}

\textbf{Fine-grained perception remains challenging for VLMs.} Existing VLMs achieve stable performance on some image-level perception tasks, but show clear performance drops on fine-grained visual understanding tasks. In vision grounding tasks such as VG, REC, and AR that require precise spatial localization, model performance remains around 50\%. For FCC that rely on fine-grained semantic distinctions, performance is generally below 50\%. For pixel-level segmentation, most models achieve only 10\%–20\% performance, and some RSVLMs approach zero on these metrics. These results are related to the large image size, multi-scale object distribution, and small-object prevalence in RSI, which increase the difficulty of fine-grained perception and localization. They also indicate that current models are mainly designed and trained for global understanding or descriptive tasks, with limited support for localization and segmentation and without task-specific architectural or training designs.

\textbf{Reasoning capability remains a major bottleneck for VLMs.} In several reasoning tasks such as SRR, FRL, GEU, and CR, general-purpose VLMs perform slightly better than RSVLMs. However, for most reasoning tasks, the performance of both general-purpose models and RSVLMs stays around or below 50\%. Models perform poorly on multi-temporal reasoning tasks, including CD and LTR. Performance is also limited on reasoning tasks that require domain knowledge, such as DAR, GL, and CR. Overall, current RSVLMs show weak reasoning performance and insufficient use of domain knowledge, indicating that reasoning and knowledge-aware modeling in remote sensing still require further study.

\textbf{VLM performance degrades significantly under low-quality and degraded image conditions.} RSI are often affected by illumination changes, such as cloud and haze, as well as noise and blur, leading to large variations in input image quality. Under degraded image conditions, the performance of most RSVLMs drops clearly. In comparison, general-purpose VLMs are more stable under blur and noise. In RGB–SAR matching tasks, most RSVLMs perform poorly, while GPT-4o maintains more consistent cross-modal matching. Overall, current RSVLMs rely heavily on their training data and generalize poorly to degraded images, long-tail cases, and unseen modalities. These results also show that modeling different modalities separately is insufficient, and that stronger alignment across modalities and resolutions is needed for cross-modal consistency.

\textbf{Current RSVLMs rely heavily on textual rather than visual information.} As shown in Figure~\ref{fig:blindtest}, model performance with text-only input is close to that obtained with both image and text inputs. Some open-source general-purpose VLMs show clear performance differences between blind and full-input settings, whereas most RSVLMs exhibit only small gaps between the two. This suggests that, despite task-specific fine-tuning on remote sensing data, RSVLM predictions in certain tasks depend only weakly on visual input. One possible explanation is that repeated training with similar task formats and textual templates encourages models to rely on textual cues or option distributions rather than image content. When the input is incomplete or ambiguous, some models further produce fixed or abnormal outputs, such as the repetitive character responses observed in GeoChat. This behavior is more evident in temporal reasoning tasks. In STR and CD, both general-purpose models and RSVLMs achieve low overall performance, with only minor differences between blind and full-input settings. Blind evaluations show that models can often generate seemingly reasonable answers based solely on textual instructions, without using image information. Overall, blind evaluation provides a practical way to assess the extent to which models rely on visual input. The results also suggest that greater attention should be given to the joint use of image and text information during model design and training, in order to reduce reliance on textual cues and improve consistency with visual evidence.

\vspace{-3pt}
\section{Conclusion}
Recent advances in vision–language models have largely been evaluated using benchmarks designed for natural images and general-domain reasoning. However, Earth observation presents challenges such as scale variation, temporal evolution, and sensing modality diversity that are not adequately captured by existing evaluations. To address this gap, we introduce OmniEarth, a systematic benchmark for evaluating vision–language models in Earth observation. OmniEarth comprises 28 fine-grained tasks in perception, reasoning, and robustness, and supports a range of vision–language problems, including recognition, localization, segmentation, counting, and temporal analysis. The benchmark includes 9,295 carefully curated images and 44,210 manually verified instructions. Through detailed experiments with 19 state-of-art VLMs, we show that existing models still exhibit limitations in fine-grained understanding, reasoning, and robustness. Overall, OmniEarth enables a more accurate assessment of current VLMs and reveals persistent gaps that remain in adapting these models to complex remote sensing applications.

\bibliographystyle{ACM-Reference-Format}
\bibliography{reference}

\newpage

\appendix

\section{Dataset Construction}\label{suppl:data}
The construction of OmniEarth aims to bridge the critical gaps in existing remote sensing benchmarks regarding multi-modality, multi-granularity, and high-order reasoning capabilities. To this end, we employed a dual strategy that combines the large-scale aggregation of existing data with the high-fidelity reconstruction of core datasets. This approach ensures that the resulting benchmark exhibits significant superiority in terms of scale, quality, and heterogeneity.

\subsection{Data Sources and Multi-modal Fusion}
The data infrastructure of OmniEarth is built upon two core pillars, designed to maximize task coverage and ensure data diversity and heterogeneity.

\noindent\textbf{Integration of Existing Datasets:} We meticulously curated and integrated 39 high-quality open-source remote sensing datasets from the community. Spanning diverse task types and sensor modalities, these datasets serve as a robust bedrock for the benchmark. Table S1 summarizes the detailed sensor attributes of these integrated datasets, including Ground Sampling Distance (GSD) and typical image resolutions. As shown in Table \ref{tab:integrated_datasets}, the details of the integrated source datasets are provided.

\noindent\textbf{Customized Acquisition for New Tasks:} To support the 28 proposed fine-grained (Level-3) tasks, we leveraged multi-source remote sensing platforms for targeted data acquisition. Subsequently, we employed hybrid annotation strategies—incorporating tag-driven methods, foundation model assistance, and Human-AI (GPT-4o) collaboration to achieve high-precision fine-grained labeling. This ensures the domain expertise and logical complexity of the generated vision-language instruction pairs.

\subsection{SAR Data Processing and Visualization}
\label{subsec:sar_processing}

The raw data acquired from the Capella Space platform is in Single Look Complex (SLC) format, denoted as $\mathbf{S} \in \mathbb{C}^{M \times N}$, where $M$ and $N$ represent the azimuth and range dimensions, respectively. Unlike optical imagery, SLC data contains both amplitude and phase information, characterized by an extremely high dynamic range and inherent speckle noise. To render this data compatible with standard vision-language models (which typically accept 8-bit RGB or grayscale inputs), we implemented a rigorous radiometric calibration and quantization pipeline. The detailed processing procedure is illustrated in Algorithm \ref{alg:sar_proc}. Figure~\ref{fig:sar-rgb} shows examples of paired SAR and optical RGB data. The pipeline consists of four critical stages designed to maximize visual interpretability while preserving geometric fidelity:

\begin{figure}[!t]
    \centering
    \includegraphics[width=\linewidth]{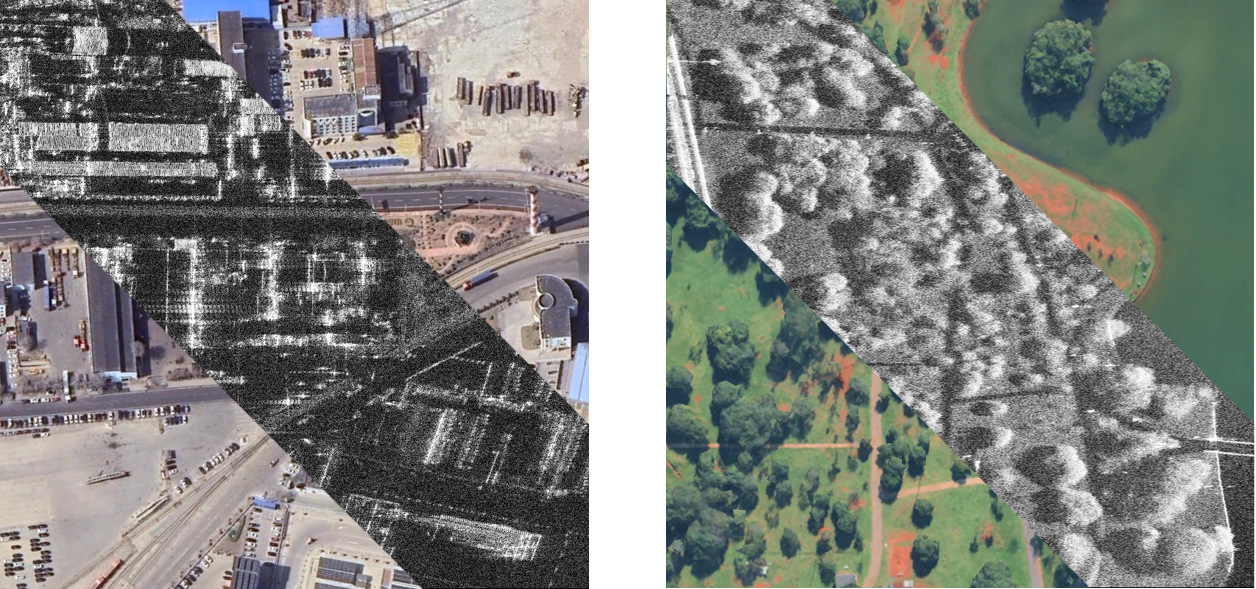}
    \caption{Examples of Corresponding SAR and RGB.}
    \label{fig:sar-rgb}
\end{figure}

\begin{algorithm}[!t]
    \caption{Processing Pipeline for Capella SLC to PNG Visualization}
    \label{alg:sar_proc}
    \SetKwInOut{Input}{Input}
    \SetKwInOut{Output}{Output}
    \SetKwFunction{Percentile}{Percentile}
    \SetKwFunction{Clip}{Clip}

    \Input{Capella SLC Data $\mathbf{S} \in \mathbb{C}^{M \times N}$ (where $M=12775, N=7346$)}
    \Output{8-bit Quantized Image $\mathbf{I}_{png} \in \mathbb{Z}^{M \times N} \in [0, 255]$}
    
    \BlankLine
    \tcp{Step 1: Amplitude Extraction}
    $\mathbf{A} \leftarrow \sqrt{\text{Re}(\mathbf{S})^2 + \text{Im}(\mathbf{S})^2}$ \tcp*{Compute magnitude from complex signal}
    
    \BlankLine
    \tcp{Step 2: Logarithmic Transformation}
    $\mathbf{P}_{dB} \leftarrow 10 \cdot \log_{10}(\mathbf{A}^2 + \epsilon)$ \tcp*{Convert to dB scale, $\epsilon=10^{-10}$}
    
    \BlankLine
    \tcp{Step 3: Contrast Enhancement (Robust Scaling)}
    $V_{min} \leftarrow \Percentile(\mathbf{P}_{dB}, 2\%)$ \tcp*{Lower bound threshold}
    $V_{max} \leftarrow \Percentile(\mathbf{P}_{dB}, 98\%)$ \tcp*{Upper bound threshold}
    $\mathbf{P}_{clipped} \leftarrow \Clip(\mathbf{P}_{dB}, V_{min}, V_{max})$ \tcp*{Outlier removal}
    
    \BlankLine
    \tcp{Step 4: Linear Quantization}
    \For{each pixel $(i, j)$ in $\mathbf{P}_{clipped}$}{
        $\mathbf{I}_{png}(i, j) \leftarrow \left\lfloor 255 \cdot \frac{\mathbf{P}_{clipped}(i, j) - V_{min}}{V_{max} - V_{min}} \right\rfloor$ \;
    }
    
    \Return $\mathbf{I}_{png}$ \;
\end{algorithm}

\textbf{(1) Amplitude Extraction:} The complex-valued SLC data is first converted into amplitude $\mathbf{A}$. This step discards the phase information, which is generally stochastic in single-frame SAR interpretation, focusing instead on the backscatter intensity.
    
\textbf{(2) Dynamic Range Compression:} SAR backscatter coefficients typically cover a vast dynamic range. Consequently, linear visualization often results in images where high-intensity features, including metal structures, overshadow low-reflectivity regions such as water or shadows. We therefore employ a logarithmic transformation to convert intensity to the decibel (dB) scale, adding a small constant $\epsilon$ to maintain numerical stability.
    
\textbf{(3) Statistical Outlier Removal:} To mitigate the impact of extreme outliers caused by corner reflectors or coherent speckle noise, we employ a percentile-based clipping strategy. We statistically determine the effective dynamic range by calculating the $2^{nd}$ and $98^{th}$ percentiles ($V_{min}$ and $V_{max}$) of the histogram. Values outside this interval are clipped, ensuring that the quantization range focuses on the majority of the scene's information.
    
\textbf{(4) Linear Quantization:} Finally, the clipped floating-point data is linearly mapped to the integer range $[0, 255]$. This generates a standard 8-bit grayscale PNG image $\mathbf{I}_{png}$, which serves as the direct visual input for the subsequent VLM annotation and evaluation tasks.

\begin{table}[t]
\centering
\caption{\textbf{Details of Integrated Source Datasets.} }
\label{tab:integrated_datasets}
\resizebox{\linewidth}{!}{
\begin{tabular}{llcc}
    \toprule
    \textbf{Dataset Name} & \textbf{Primary Task} & \textbf{Modality} & \textbf{GSD} \\
    \midrule

    \multicolumn{4}{l}{\textit{\textbf{SAR Datasets}}} \\
    AIR-SARship-1 & OC, VG, IMR, SC & SAR & 1.3\,m \\
    AIR-SARship-2 & OC, VG, IMR, SC & SAR & 1.3\,m \\
    SAR-Aircraft & OC, VG, IMR, SC & SAR & 1\,m \\
    ShipDataset & OC, VG & SAR & $3 \sim 25$\,m \\
    FIAR & OC, VG & SAR & $1 \sim 5$\,m \\
    SRSDD & IC, VG, REC & SAR & 1\,m \\
    FuSAR-Ship & IC, VG, REC & SAR & $1 \sim 2$\,m \\
    FAIR-CSAR & IC, VG, REC & SAR & $1 \sim 5$\,m \\
    SRSDD-V1.0 & IC, VG, REC & SAR & 1\,m \\
    SARDet-100k & IC, VG, REC & SAR & N/A \\
    BRIGHT & DAR, DCI & SAR & $0.3 \sim 1$\,m \\
    SOS & RES & SAR & 10\,m \\
    Capella Space & LCC, REC, IC, SEC & SAR & N/A \\

    \midrule
    \multicolumn{4}{l}{\textit{\textbf{Multi-Spectral \& Hybrid Datasets}}} \\
    TreeSatAI & IMR & MS & 0.2\,m \\
    Atlantic & IMR & MS & 10\,m \\
    C2Seg-BW & IMR & MS & 10\,m \\
    SATLAS & PS, IC & RGB/MS & 1\,m \\

    \midrule
    \multicolumn{4}{l}{\textit{\textbf{RGB Datasets}}} \\
    fMow & SC, DVQA & RGB & 1\,m \\
    AID & LCC & RGB & $0.3 \sim 1$\,m \\
    JL-1 & SC, FCC, DAR, STR, CR & RGB & 0.5\,m \\
    DMSP-OLS & IMR & Night & N/A \\
    DIOR & IC, VG, REC & RGB & N/A \\
    DOTA & IC, VG, DVQA & RGB & $0.3 \sim 1$\,m \\
    AIRS & IC, DVQA & RGB & 0.075\,m \\
    MAR20 & FCC & RGB & N/A \\
    SODA & OC, DVQA & RGB & N/A \\
    FAIR1M2.0 & FCC & RGB & $0.3 \sim 0.8$\,m \\
    FGSC-23 & FCC & RGB & $0.4 \sim 2$\,m \\
    iSAID & RES, GRES & RGB & N/A \\
    LoveD & RES, GRES & RGB & 0.3\,m \\
    SECOND & CD & RGB & N/A \\
    JL1-CD & CD, CMS & RGB & $0.5 \sim 0.75$\,m \\
    TGRS-HRRSD & AR, SRR & RGB & $0.15 \sim 1.2$\,m \\
    WHU Building & \raggedright AR, GM, DAR, DCI, ICA, HD, FRL & RGB & 0.2\,m \\
    DIUx xView & ICA, HD, FRL & RGB & 0.3\,m \\
    WHU Cloud & ICA & RGB & 30\,m \\
    LEVIR & ICA & RGB & $0.2 \sim 1$\,m \\
    Flood-3i & Flood RES, GM, DC & RGB & 0.05\,m \\
    xBD & FRL, CD, DSR, DCI & RGB & 0.3\,m \\
    CD\_Data\_GZ & CD, CMS & RGB & 0.55\,m \\
    EBD & FRL, CD, DAR, DCI & RGB & 0.3\,m \\
    Google Earth & LCC, LTR, GL, GEU & RGB & N/A \\
    World Imagery & LCC, LTR, GL, GEU & RGB & N/A \\
    \bottomrule
\end{tabular}}
\end{table}

\section{Definitions of Fine-Grained Tasks}\label{suppl:tasks}
\begin{figure}[h]
  \centering
  \begin{minipage}{0.48\linewidth}
    \centering
    \includegraphics[width=\linewidth]{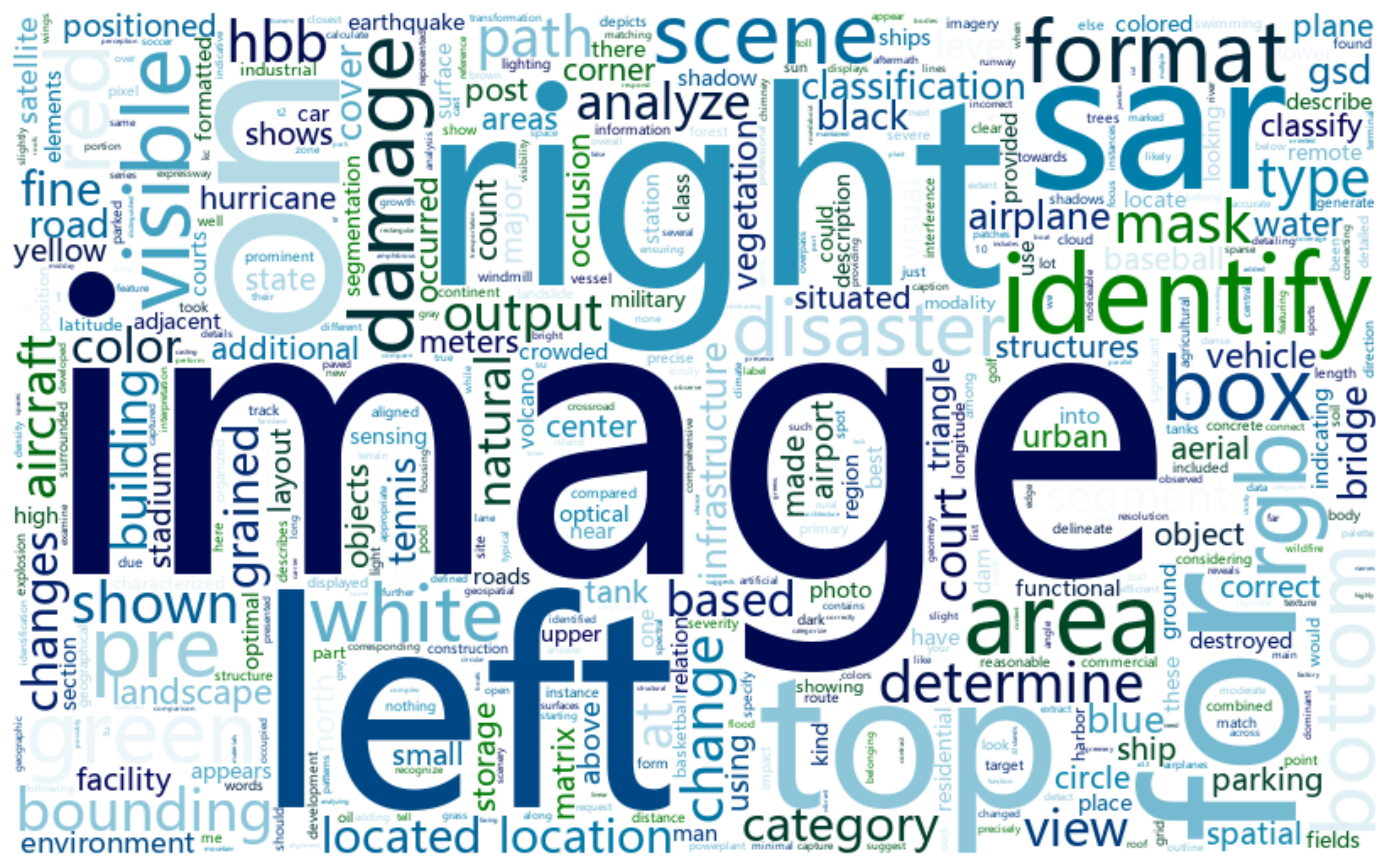}\\
    {\small (a) Overall}
  \end{minipage}\hfill
  \begin{minipage}{0.48\linewidth}
    \centering
    \includegraphics[width=\linewidth]{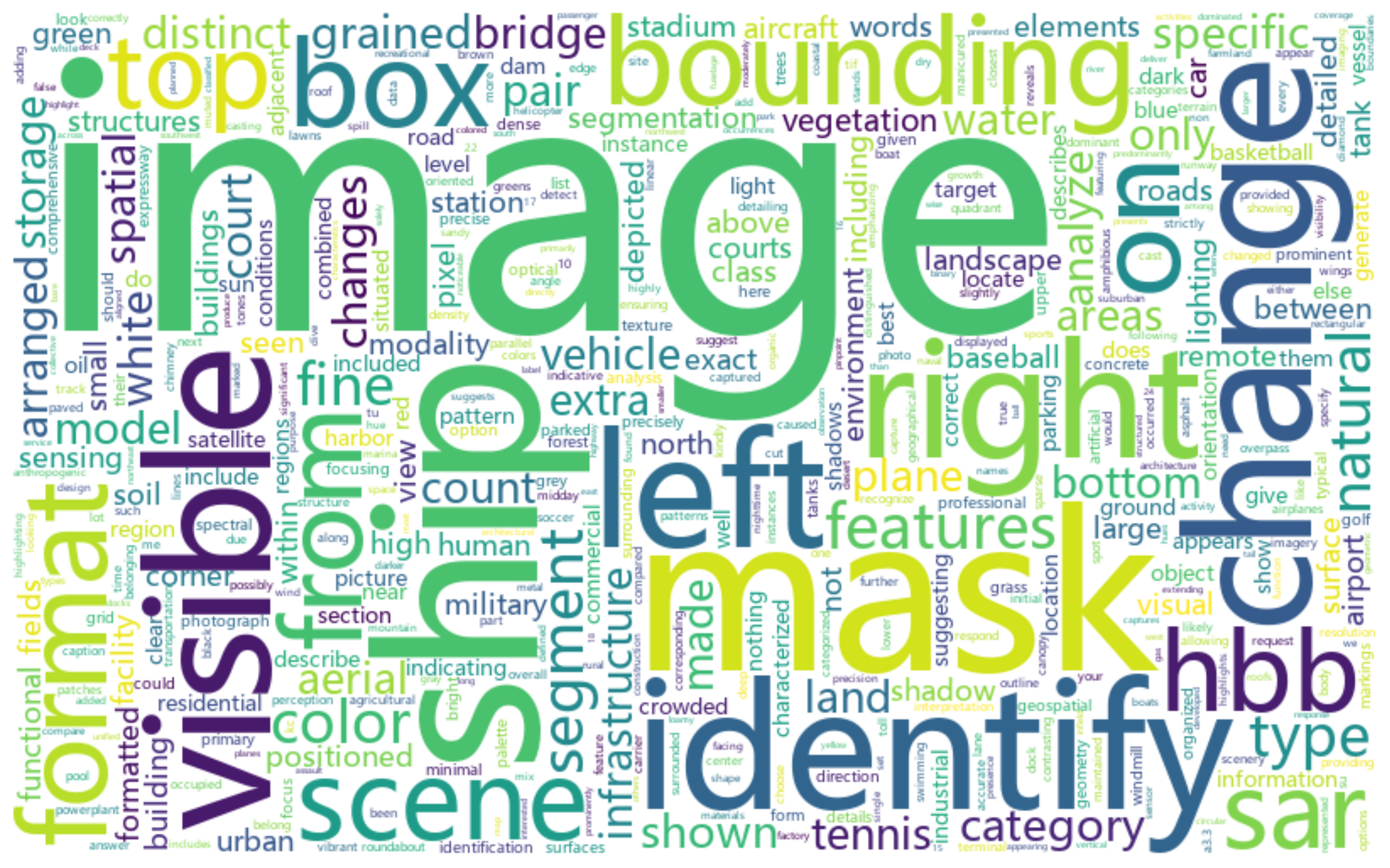}\\
    {\small (b) Perception}
  \end{minipage}

  \vspace{0.3em}

  \begin{minipage}{0.48\linewidth}
    \centering
    \includegraphics[width=\linewidth]{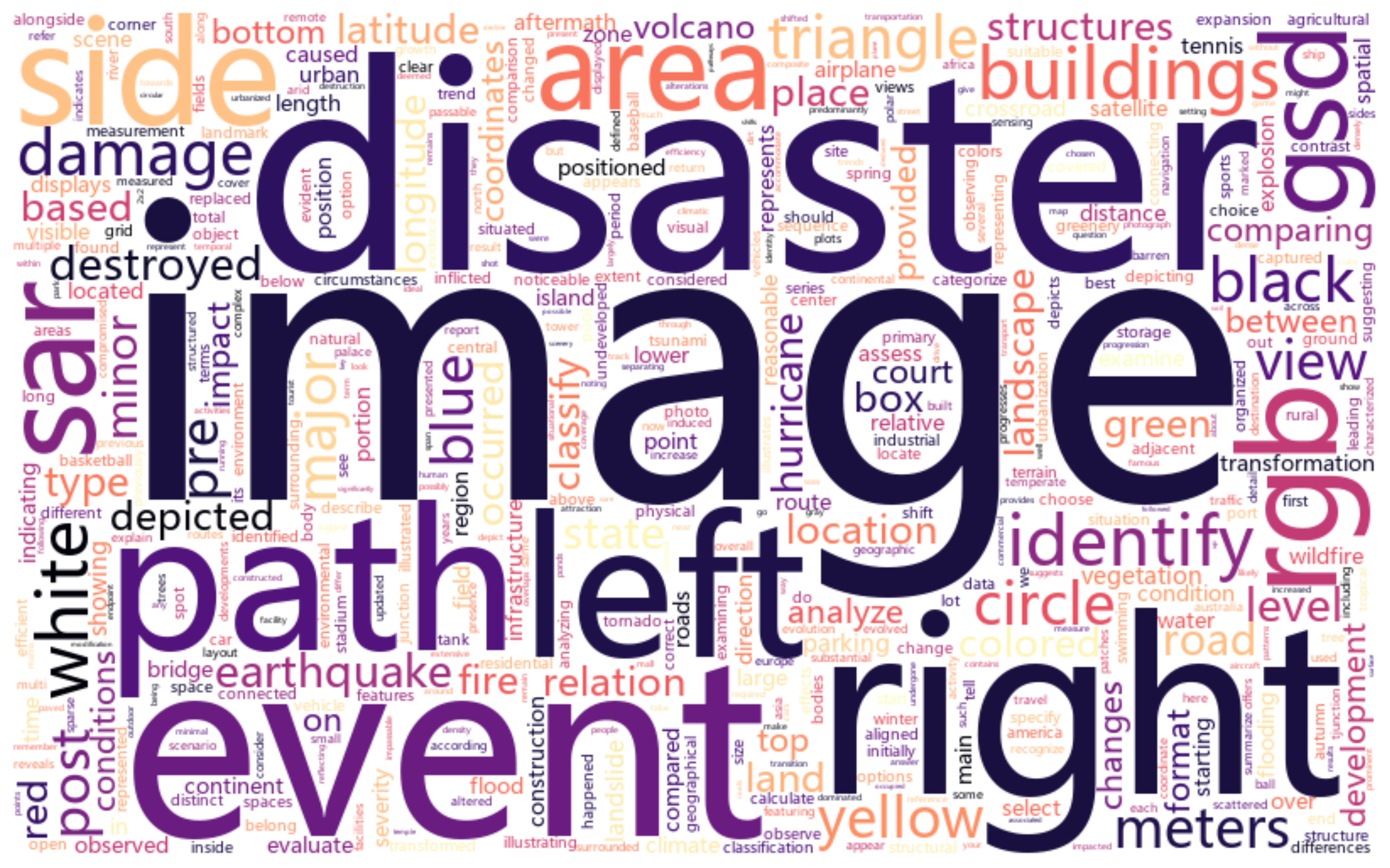}\\
    {\small (c) Reasoning}
  \end{minipage}\hfill
  \begin{minipage}{0.48\linewidth}
    \centering
    \includegraphics[width=\linewidth]{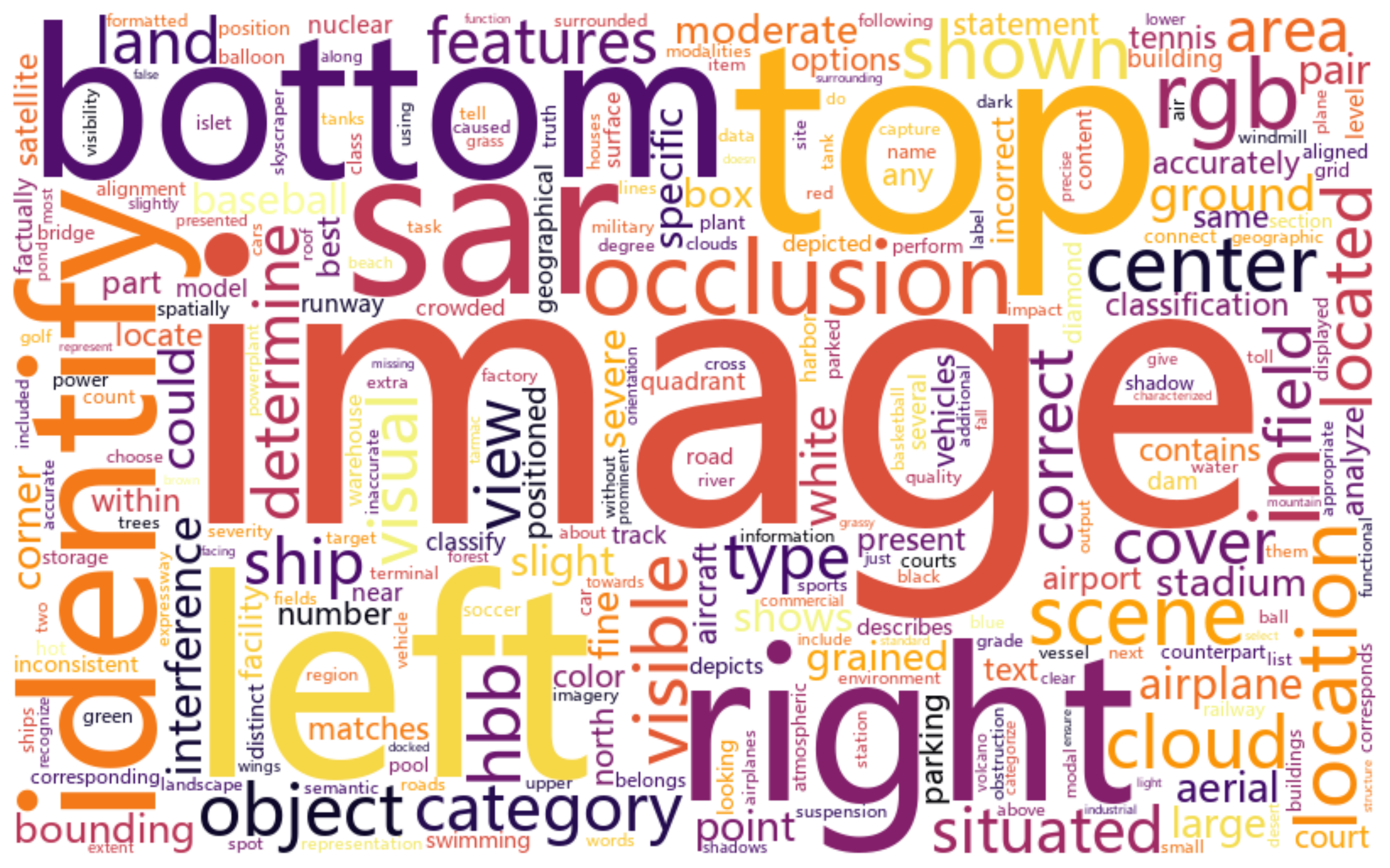}\\
    {\small (d) Robustness}
  \end{minipage}

  \caption{Word clouds of OmniEarth and its three evaluation dimensions.}
  \label{fig:wordcloud-2x2}
\end{figure}

OmniEarth provides a comprehensive framework to assess Vision-Language Models (VLMs) in remote sensing and geospatial analysis. It adopts a hierarchical structure centered on three core dimensions such as Perception, Reasoning, and Robustness. These are organized into 8 sub-categories and cover 28 fine-grained geospatial tasks. To intuitively reflect the task characteristics and keyword distribution of each core dimension as well as the overall benchmark, the word cloud diagrams corresponding to different tasks are shown in Figure \ref{fig:wordcloud-2x2}.

\subsection{Image-Level Perception}
As the primary sub-task category under the Perception dimension of the OmniEarth, Image-level Understanding focuses on the high-level, synoptic interpretation of global semantics and inherent attributes within RSI. This category comprises four fine-grained tasks designed to comprehensively evaluate the macro-cognitive capabilities of Vision-Language Models (VLMs) when navigating complex geospatial scenes.

\textbf{Scene Classification (SC):} 
As a basic task in RSI interpretation, SC assesses a model’s ability to understand high-level semantics and distinguish different scene types. The task requires models to integrate spectral, textural, and spatial features to categorize common anthropogenic scenes, such as airports and harbors. SC is an important component of large-scale land use and land cover analysis. The data are collected from the fMoW dataset, proprietary JL-1 imagery, and manually cropped samples from Capella Space Open Data. The dataset includes 260 RGB images and 140 SAR images, covering 75 artificial scene categories, including aircraft, vehicles, bridges, roads, and buildings.

\textbf{Land-Cover Classification (LCC):} This task targets the classification of natural terrestrial surfaces. It requires models to use global image characteristics to distinguish common land-cover types, such as bare land, forests, oceans, deserts, rainforests, and lakes. By covering a range of natural categories with different spectral and textural properties, the task evaluates a model’s ability to recognize diverse natural landscapes.  The data are collected from the AID dataset, Esri World Imagery Wayback, and Capella Space Open Data. The dataset contains 178 RGB images and 22 SAR images, covering 15 natural scene categories, including bare land, beaches, deserts, farmlands, forests, and meadows.

\textbf{Image Modality Recognition (IMR):} This task focuses on identifying the sensing modality of a given RSI. It evaluates whether a model can distinguish between different image types based on their imaging characteristics, including RGB optical imagery, SAR, multispectral imagery, and nighttime light data.  We consider four modalities: RGB, SAR, Nighttime Light (NIGHT), and False Color. RGB images are collected from Flood-3i. False Color images are generated by randomly selecting non-standard spectral bands from C2Seg-BW, Atlantic, and TreeSatAI-Time-Series. SAR data are sourced from the SOS dataset, and NIGHT images are taken from DMSP-OLS. In total, 140 MCQs are constructed for modality recognition.

\textbf{Image Captioning (IC):} As a common task in vision–language evaluation, IC assesses a model’s ability to generate natural language descriptions for RSI. The captions are expected to summarize the main scene content, including geospatial context, key objects, and overall spatial layout. The dataset contains 250 RGB images and 50 SAR images, covering a range of scenes from rural areas and ports to urban environments. All question–answer pairs are produced using a standardized pipeline with GPT-4o generation followed by human verification.

\subsection{Object-level Perception}
This dimension shifts the focus from global scene understanding to the recognition and localization of specific entities and regions. It emphasizes identifying and describing local geospatial objects without relying on pre-defined category assumptions.

\textbf{Visual Grounding (VG):} This task evaluates a model’s ability to localize a target object in an image based on a natural language description. Given a referring expression, the model is required to identify a single target instance and output its corresponding horizontal bounding box (HBB). The task focuses on aligning language with local visual regions while handling background clutter and small objects in RSI. The RGB data include 800 images from the DIOR and DOTA datasets, and the SAR data include 200 images from MAR20. All samples are manually filtered to remove inaccurate annotations, and the question formats and referring expressions are carefully redesigned to improve clarity and consistency.

\textbf{Referring Expression Comprehension (REC):} This task extends object localization to scenarios involving multiple target instances. Given a referring expression, the model is required to identify all objects that satisfy the described conditions. The expressions may involve spatial relations, comparisons, or attribute constraints, requiring the model to reason over relationships among multiple objects. The dataset includes 399 RGB images from DIOR and 50 SAR images from MAR20. All referring expressions and corresponding object annotations are carefully verified by human experts.

\textbf{Object Counting (OC):} This task evaluates a model’s ability to count target objects within a scene. It requires handling challenges such as object overlap, occlusion, and scale variation between the global scene and local regions of interest. In addition to general object counting, the task also includes counting damaged buildings in post-disaster scenarios, which involves irregular object shapes. Initial counts for both RGB and SAR images are generated by GPT-4o and then reviewed and corrected by human annotators to produce the final question–answer pairs.

\textbf{Fine-grained Category Classification (FCC):} This task focuses on distinguishing visually similar object categories with subtle intra-class differences. Models are required to recognize fine-grained structural features to correctly classify objects within the same general category. For example, distinguishing aircraft models such as Boeing 747 and Y-20 relies on differences in wing structure and engine configuration. The data are collected from FAIR1M2.0, FGSC-23, and proprietary JL-1 imagery, comprising 365 images in total. All images and questions are manually reviewed to ensure annotation accuracy.

\textbf{Attribute Recognition (AR):} This task focuses on identifying specific attributes of geospatial entities beyond category labels. Models are required to recognize physical properties such as orientation, size, and color, as well as scene-level attributes related to operational states, including crop growth stages and parking lot occupancy. The task involves understanding spatial patterns and temporal or environmental cues present in the imagery. A test set of 300 images is constructed with manually designed questions and answer options, covering attributes such as object color, object orientation, and scene or venue density.

\subsection{Pixel-level Perception}
This category focuses on precise discrimination and segmentation of fine-grained object boundaries and spatio-temporal change regions in RSI. Compared with object-level recognition, these tasks place higher demands on spatial resolution handling, boundary feature extraction, and the association of information across multiple time steps.

\textbf{Referring Expression Segmentation (RES):} This task evaluates a model’s ability to produce pixel-wise segmentation masks for a target region specified by a natural language description. Given a referring expression, the model is required to accurately delineate the object boundaries at the pixel level. The data are collected from iSAID and LoveD (300 RGB images) and the SOS dataset (50 SAR images). After manually removing inaccurate masks, the question formats and referring expressions are reorganized for consistency.

\textbf{Generalized Referring Expression Segmentation (GRES):} This task extends RES to more complex scenarios. It supports cases involving multiple target instances as well as situations where no valid target is present. The dataset includes 255 RGB images selected from iSAID and LoveD. All samples are manually cleaned to remove ambiguous annotations and reformatted to support generalized referring expressions.

\textbf{Change Mask Segmentation (CMS):} This task focuses on pixel-level change detection in bi-temporal RSI. Given a pair of images and a textual query, the model is required to generate a segmentation mask that highlights regions of change, capturing both the location and extent of geophysical variations. The dataset comprises 200 questions sourced from CD\_DATA\_GZ, SECOND, and proprietary JL-1 imagery. All samples are manually inspected and corrected.

\begin{figure}[!t]
    \centering
    \includegraphics[width=\linewidth]{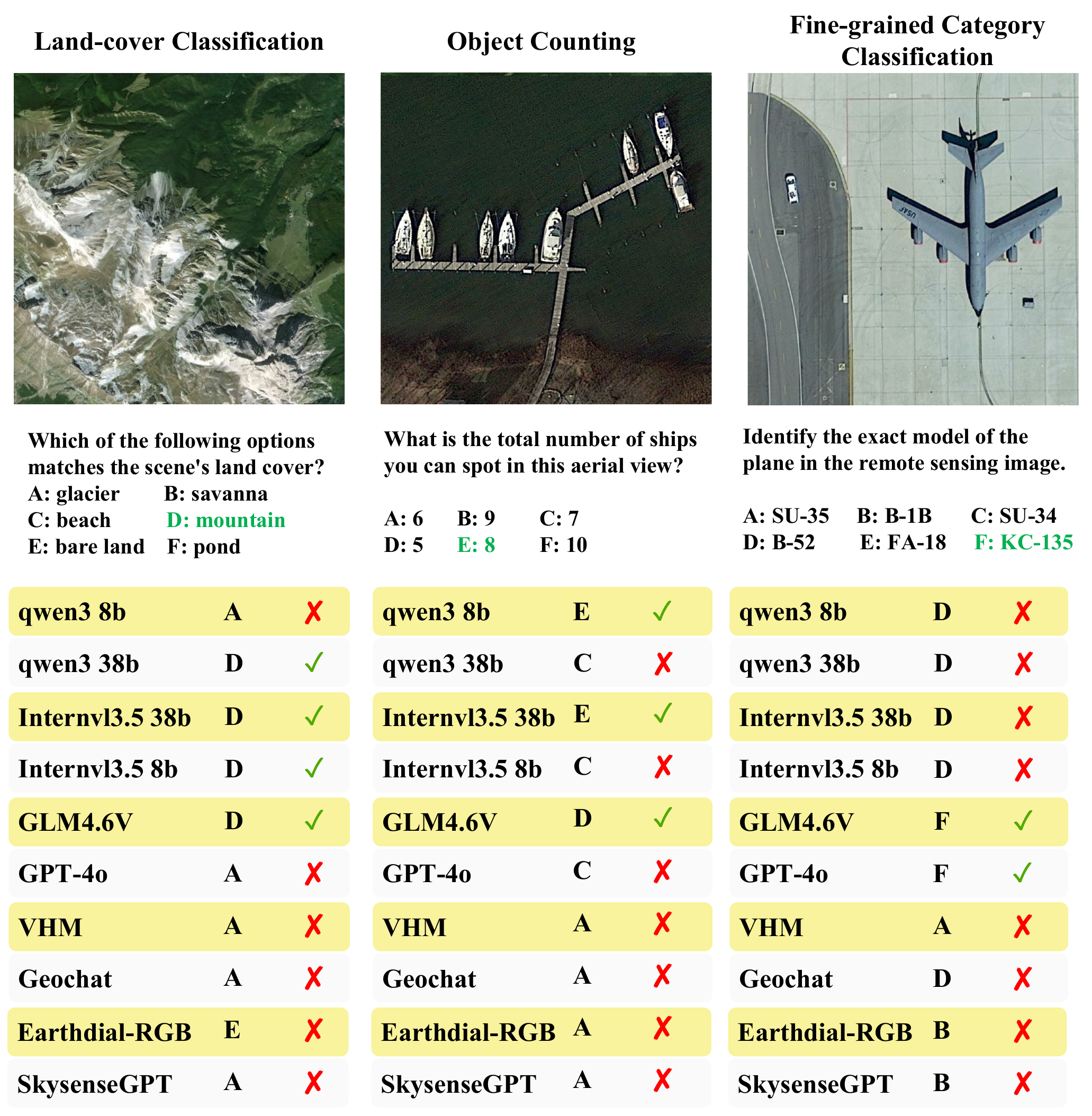}
    \caption{Sample Diagram of Perceptual Task Model Responses.}
    \label{fig:adi_per}
\end{figure}

Figure~\ref{fig:adi_per} shows representative qualitative examples from selected tasks within the three image-, object- and pixel-level categories.

\subsection{Spatial Reasoning}
This category focuses on analyzing spatial relationships between entities and the geometric properties of individual objects. The tasks require models to go beyond basic recognition by using the overhead view of RSI together with geospatial knowledge. Models are expected to interpret spatial relationships, such as orientation, distance, and containment, and to estimate geometric attributes including area, length, and distance. These capabilities are relevant to applications such as urban planning and resource management.

\textbf{Spatial Relationship Reasoning (SRR):} This task evaluates a model’s ability to interpret spatial relationships between objects in complex scenes. It involves understanding relative positions, such as left and right, as well as spatial relations including adjacency and containment. These relationships are essential for forming a structured understanding of geospatial scenes. The dataset is based on 315 images from TGRS-HRRSD. Referring expressions are manually written, after which GPT-4o generates candidate answers and distractors that are further refined by human annotators.

\textbf{Geometric Measurement (GM):} This task focuses on estimating geometric properties of objects in RSI. Models are required to measure attributes such as building footprint area, road length, plot perimeter, and distances between entities. The dataset includes 300 images with accurate segmentation masks from Flood-3i and the WHU Building Dataset. Using known ground sampling distance (GSD), questions related to physical measurements are manually constructed.

\textbf{Functional Region Localization (FRL):} This task examines whether models can identify regions based on their functional roles rather than explicit visual categories. It requires locating areas such as emergency shelter sites, irrigated farmland, or post-disaster resettlement zones by combining visual cues with contextual reasoning. The data are collected from xBD, the WHU Building Dataset, and DIUx xView, totaling 172 images. All questions and answers are manually designed by experts to assess functional region understanding.

\subsection{Temporal Reasoning}
This category focuses on analyzing changes and trends from multi-temporal RSI. Unlike single-time interpretation, these tasks require models to compare information across different time periods to identify short-term changes and longer-term evolution, such as urban expansion or vegetation variation. Such capabilities are relevant for applications including resource management and disaster assessment.

\textbf{Change Description (CD):} This task evaluates a model’s ability to describe changes between bi-temporal images. Models are required to identify the type of change, such as land-use conversion, and summarize its extent in natural language. The dataset includes 260 images selected from xBD, CD\_DATA\_GZ, SECOND, and JL1-CD. Initial descriptions are generated by GPT-4o and then manually verified and corrected for consistency and logic.

\textbf{Damage Assessment Reasoning (DAR):} This task focuses on assessing infrastructure damage by comparing pre- and post-disaster imagery. Models are required to determine damage levels for buildings and roads, distinguishing between severe and partial damage, and to assess road accessibility for emergency response. The dataset integrates 381 images from xBD, BRIGHT, the WHU Building Dataset, Flood-3i, and proprietary JL-1 imagery. The task emphasizes reasoning based on visual and geographic evidence.

\textbf{Long-term Trend Reasoning (LTR):} This task evaluates a model’s ability to summarize long-term changes from sequential imagery. It involves identifying broad trends such as urban growth, vegetation changes, and variations in water body extent. We manually curate 100 image sequences (533 images in total) from Esri World Imagery Wayback. Trend descriptions are assisted by GPT-4o and subsequently verified by human annotators.

\textbf{Seasonal Temporal Reasoning (STR):} This task assesses whether a model can infer temporal attributes such as season or time of day based on visual cues, including vegetation appearance and shadow patterns. The dataset is built using proprietary JL-1 imagery, covering 101 regions across four seasons with composite comparison images. All questions and answer options are manually designed.

\subsection{Geographic Application Reasoning}
This category focuses on reasoning about geographic attributes and practical application scenarios using RSI. It evaluates whether models can combine visual evidence with geographic knowledge to support application-oriented analysis.

\textbf{Geo-localization Reasoning (GL):} This task evaluates a model’s ability to infer geographic information such as approximate latitude and longitude ranges, continental location, and climate zones. Models are required to use visual cues including terrain structure, vegetation patterns, and land-use characteristics. The dataset consists of 237 representative images manually selected from Esri World Imagery Wayback, with corresponding geographic inference questions constructed by experts.

\textbf{Disaster Cause Inference (DCI):} This task focuses on identifying the causes of disasters based on post-event imagery. Models must infer triggers such as explosions, wildfires, floods, or typhoons by analyzing surface damage patterns and spatial distributions. The dataset integrates RGB images from xBD and SAR images from BRIGHT. Disaster types are categorized and sampled using a proportional stratified strategy, resulting in 254 RGB and 271 SAR MCQs.

\textbf{Geo-Entity Understanding (GEU):} This task evaluates the recognition of well-known geographic entities, including landmarks and island groups, based on their visual appearance and surrounding spatial context. The dataset includes 150 images manually collected from Esri World Imagery Wayback, covering 50 islands and 100 landmarks, and is designed to test geographic entity recognition.

\textbf{City Recognition (CR):} This task examines whether models can identify specific cities from RSI. It requires extracting large-scale spatial patterns such as urban layout, road network structure, and functional zoning. The dataset contains 322 images of different cities collected from proprietary JL-1 imagery, with questions designed to assess city-level identification.

\textbf{Planning Suggestions (PS):} This task evaluates a model’s ability to reason about planning-related decisions under geographic constraints. Given multiple candidate routes, models are required to select suitable options by considering factors such as terrain relief, road connectivity, and obstacles. The dataset is based on SATLAS, where four candidate routes are manually designed for each scenario to test reasoning over road topology and terrain conditions.

\begin{figure}[!t]
    \centering
    \includegraphics[width=\linewidth]{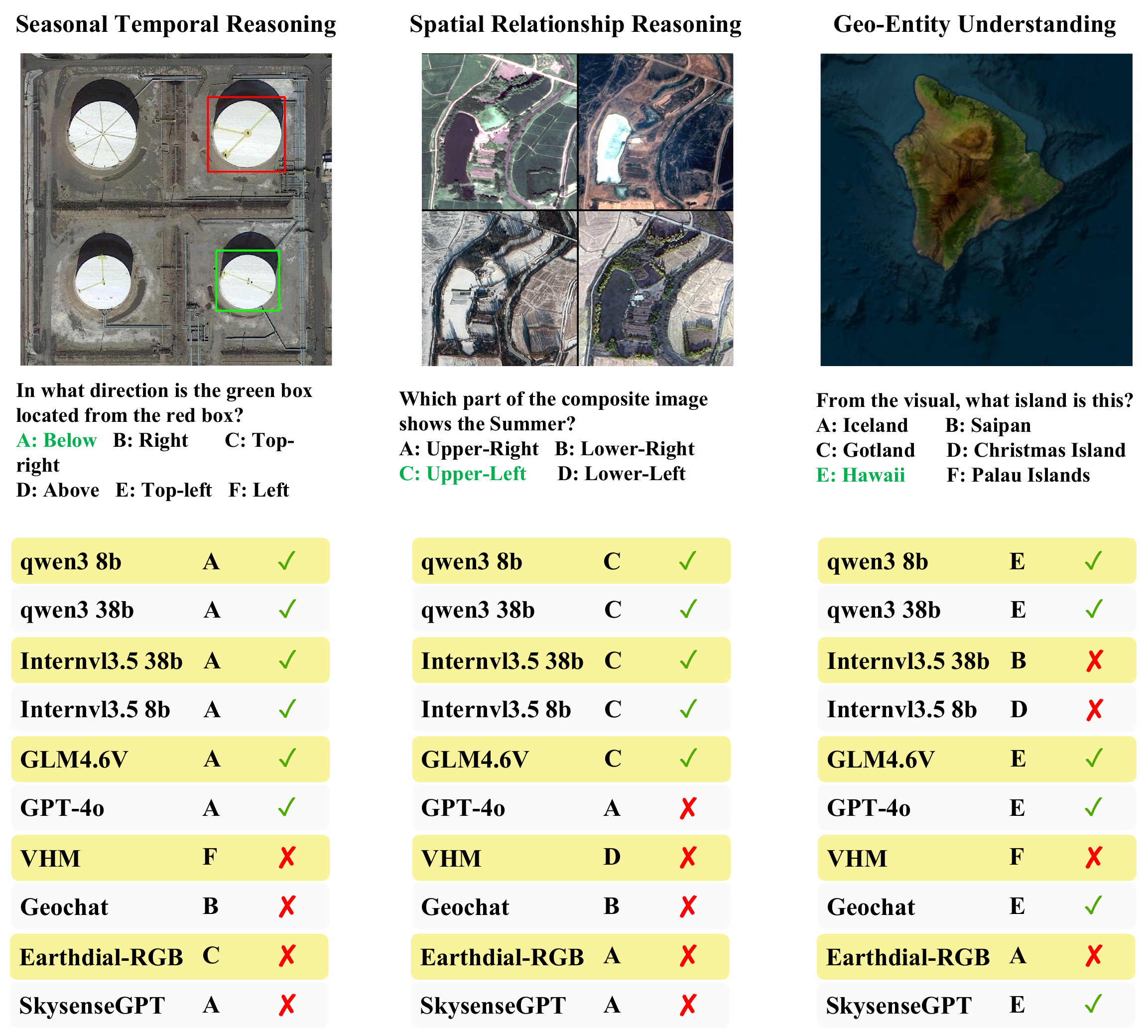}
    \caption{Sample Diagram of Reasoning Task Model Responses.}
    \label{fig:adi_res}
\end{figure}

The complexity of high-level geospatial cognition is further illustrated in Figure~\ref{fig:adi_res}.

\subsection{Environmental Resilience}
This category evaluates how models perform under common environmental interference and data degradation in RSI. It focuses on the stability of perception and reasoning when images are affected by factors such as cloud cover, occlusion, geometric distortion, or reduced resolution, which are typical in real-world acquisition scenarios.

\noindent\textbf{Image Condition Assessment (ICA):} This task evaluates a model’s ability to assess image usability and identify environmental interference, including clouds, snow, and shadows. By detecting these factors, the model can provide useful quality cues for downstream analysis. The dataset consists of 200 images manually selected from the WHU Building Dataset, WHU Cloud, and DIUx xView, designed to test judgments of image quality and environmental conditions.

\noindent\textbf{Degraded-condition VQA (DVQA):} This task examines the robustness of vision–language reasoning under image degradation. It includes geometric transformations, such as rotation and affine distortion, as well as resolution reduction. The task evaluates whether models can maintain consistent semantic understanding when image quality is reduced.

\subsection{Semantic Reliability}
This category focuses on the correctness and consistency of model-generated semantics. It evaluates whether models can avoid producing unsupported information and maintain stable semantic understanding across different sensing modalities.

\textbf{Hallucination Detection (HD):} This task assesses whether model outputs are grounded in the visual evidence. Given multiple candidate descriptions, the model is required to identify which options are consistent with the image content. The task is formulated as MCQs with one to four correct answers, encouraging careful verification and reducing reliance on unsupported assumptions. To benchmark model robustness against non-ideal observation conditions, we systematically simulate nine types of degradations covering the full remote sensing acquisition pipeline. Let $\mathbf{I}_{clean}$ denote the original image and $\mathbf{I}_{deg}$ the degraded output. The specific mathematical formulations are categorized as follows: \textbf{(1) Sensor Noise \& Filtering:} We model sensor-induced artifacts using additive noise and convolutional distortions. \textit{Gaussian \& Impulse Noise:} We apply Additive White Gaussian Noise (AWGN) $\mathbf{I}_{deg} = \mathbf{I}_{clean} + \mathcal{N}(0, \sigma^2)$ to simulate thermal noise, and Salt-and-Pepper noise (replacing pixels with $0$ or $255$ with probability $p$) to simulate bit errors. \textit{Blurring:} We model optical defocus and platform jitter via convolution: $\mathbf{I}_{deg} = \mathbf{I}_{clean} * k$, where $k$ represents either an isotropic Gaussian kernel $G_{\sigma}$ or a linear motion kernel $M_{\theta, l}$. \textbf{(2) Atmospheric \& Photometric Perturbations:} We simulate environmental interferences based on physical imaging models. \textit{Haze:} Modeled by the Atmospheric Scattering Equation: $\mathbf{I}_{deg}(\mathbf{x}) = \mathbf{I}_{clean}(\mathbf{x})t(\mathbf{x}) + A(1-t(\mathbf{x}))$, where $A$ is the global atmospheric light and $t(\mathbf{x})$ is the transmission map. \textit{Cloud Occlusion:} Simulated via alpha blending with synthesized cloud patterns $\mathbf{C}$: $\mathbf{I}_{deg} = \alpha \mathbf{I}_{clean} + (1-\alpha)\mathbf{C}$. \textit{Brightness \& Contrast:} Modeled as a linear transformation $\mathbf{I}_{deg} = a \mathbf{I}_{clean} + b$, simulating varying solar illumination angles and exposure settings. \textbf{(4)Transmission \& Storage Defects:} We replicate artifacts arising from data compression and sensor failure. \textit{Compression Artifacts:} We apply JPEG-like distortion via discrete cosine transform (DCT) and quantization $\mathcal{Q}$: $\mathbf{I}_{deg} = \text{IDCT}(\mathcal{Q}(\text{DCT}(\mathbf{I}_{clean})))$. \textit{Data Gaps:} We simulate sensor line dropouts (e.g., SLC-off) by applying a binary mask $\mathbf{M}$: $\mathbf{I}_{deg} = \mathbf{I}_{clean} \odot \mathbf{M}$, where masked regions are set to zero.

\textbf{Semantic Consistency (SE):} This task evaluates cross-modal semantic alignment between different sensors. Given an RGB image as a query, the model must select the corresponding geographic region from several SAR candidates. The dataset contains 300 groups, each consisting of one RGB query and three SAR candidates, and includes no-match cases to test the model’s ability to correctly reject incorrect options.

\begin{figure}[!t]
    \centering
    \includegraphics[width=\linewidth]{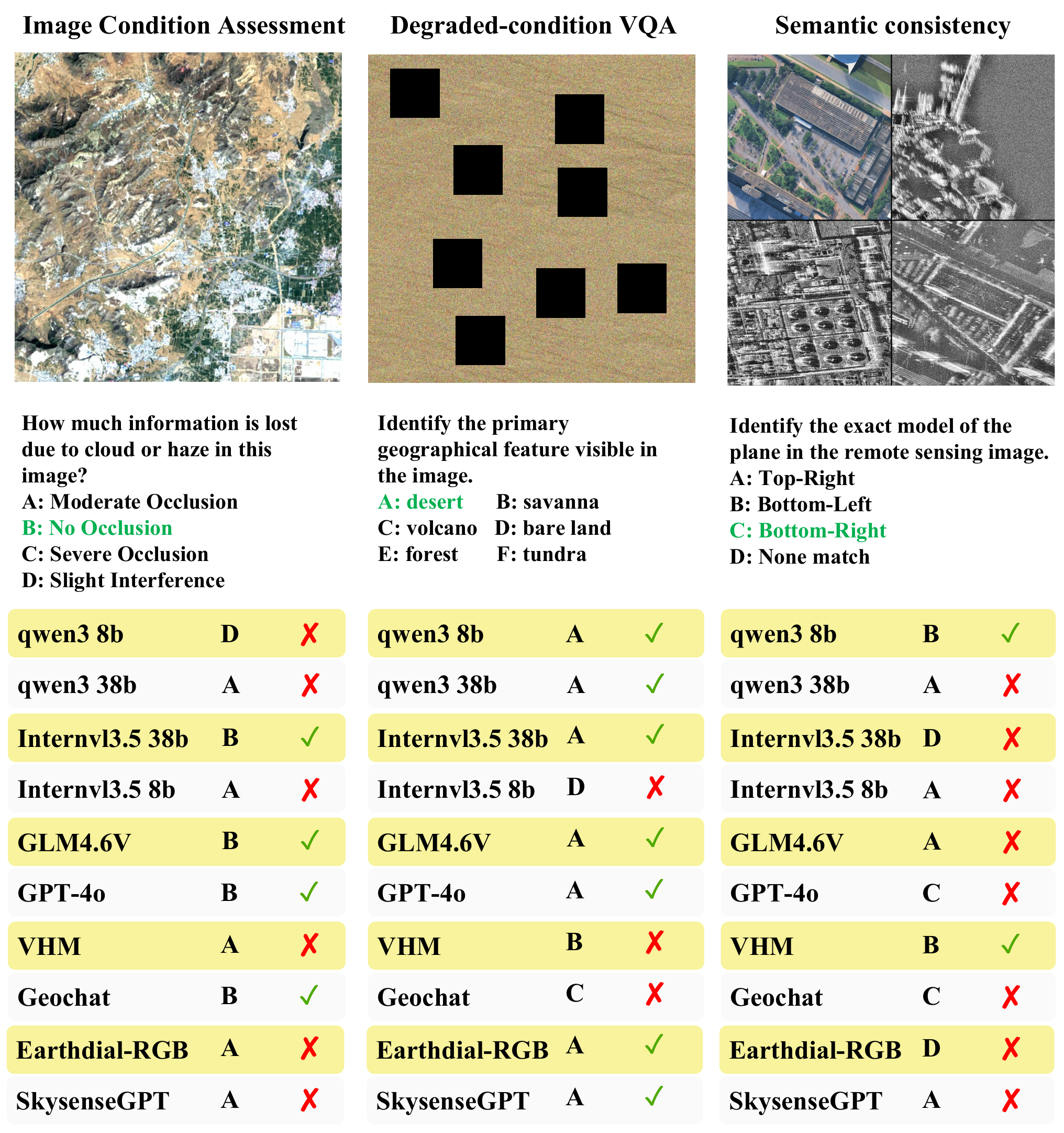}
    \caption{Sample Diagram of Robustness Task Model Responses.}
    \label{fig:adi_rob}
\end{figure}

Figure~\ref{fig:adi_rob} illustrates the models' reliability and resilience under challenging conditions. 

\section{Experimental Results}
\label{suppl:exp}

This section provides the complete experimental tables omitted from the main paper (Tabel~\ref{tab:det}, Tabel~\ref{tab:det_delta}, Tabel~\ref{tab:gen_base} and Tabel~\ref{tab:gen_delta}).

Table~\ref{tab:det} lists detailed indicators for detection tasks including visual grounding and functional region localization. Qwen3-VL-8B reaches an Acc@0.5 of 64.8 in visual grounding, outperforming the larger Qwen2.5-VL-72B. InternVL3.5-8B shows improved performance compared to InternVL3-8B across visual grounding and referring expression comprehension, with visual grounding Acc@0.5 increasing from 19.6 to 53.1. 

Table~\ref{tab:det_delta} analyzes the visual gain for detection tasks using delta metrics. General models exhibit higher delta values in recognition, with Qwen3-VL-8B achieving 63.3 in visual grounding. Remote sensing specialized models show lower gain in these tasks, such as GeoChat with a visual grounding delta of 1.4.

Table~\ref{tab:gen_base} lists base metrics for generative tasks including image captioning, change description, and long term trend reasoning. Proprietary models GPT-4o and Gemini-2.0-Flash achieve high CIDEr scores in image captioning and trend reasoning. Specialized models like GeoChat and VHM-7B also show competitive performance in descriptive tasks, with CIDEr scores reaching 138.1 and 113.0 respectively.

Table~\ref{tab:gen_delta} displays the visual gain for generative tasks. Although some specialized models obtain high base scores, their delta metrics remain low, such as VHM-7B with a CIDEr delta of 0.9 in image captioning, indicating that the generated content is weakly linked to visual evidence.

---

\begin{table*}[!h]
    \caption{Full Evaluation of Detection Tasks}
    \label{tab:det}
    \centering
    \resizebox{\linewidth}{!}{
        \begin{tabular}{l | cccc | cccc | cccc}
            \toprule
            \multirow{2}{*}{\textbf{Method}} & \multicolumn{4}{c|}{\textbf{A2.1 Visual Grounding}} & \multicolumn{4}{c|}{\textbf{A2.2  Referring Expression Comprehension}} & \multicolumn{4}{c}{\textbf{B1.3 Functional Region Localization}} \\
            \cmidrule(lr){2-5} \cmidrule(lr){6-9} \cmidrule(lr){10-13}
            & mIoU & Acc@0.25 & Acc@0.5 & Acc@0.75 & mIoU & Acc@0.25 & Acc@0.5 & Acc@0.75 & mIoU & Acc@0.25 & Acc@0.5 & Acc@0.75 \\
            \midrule
            \rowcolor{gray!10} \multicolumn{13}{l}{\textit{General Close-source VLMs}} \\
            GPT-4o & 20.7 & 39.4 & 16.6 & 2.0 & 10.1 & 14.7 & 3.6 & 0.00 & 21.8 & 38.9 & 13.4 & 0.6 \\
            Claude-sonnet-4 & 24.4 & 44.8 & 14.1 & 1.4 & 10.3 & 15.8 & 2.0 & 0.0 & 27.5 & 46.5 & 28.5 & 6.4 \\
            Gemini-2.0-Flash & 35.2 & 64.7 & 30.1 & 3.9 & 19.9 & 33.6 & 11.8 & 0.89 & 26.0 & 47.1 & 19.2 & 3.5 \\
            \midrule
            \rowcolor{gray!10} \multicolumn{13}{l}{\textit{General Open-source VLMs}} \\
            GLM-4.6V & 56.3 & 73.9 & 60.2 & 41.2 & 56.3 & 68.6 & 63.0 & 47.9 & 53.4 & 72.7 & 62.2 & 37.8 \\
            InternVL3.5-38B & 59.7 & 78.9 & 59.8 & 47.6 & 12.3 & 16.9 & 13.8 & 7.6 & 35.1 & 51.2 & 38.4 & 21.5 \\
            InternVL3.5-8B & 45.8 & 64.7 & 53.1 & 29.9 & 15.1 & 20.7 & 18.9 & 8.8 & 29.1 & 45.4 & 29.1 & 12.2 \\
            InternVL3-8B & 24.0 & 39.9 & 19.6 & 6.7 & 10.7 & 19.6 & 3.6 & 0.2 & 13.7 & 26.2 & 2.9 & 0.6 \\
            Qwen3-VL-32B & 52.9 & 70.9 & 56.1 & 38.2 & 40.0 & 55.0 & 41.0 & 25.2 & 58.5 & 83.1 & 69.8 & 36.6 \\
            Qwen3-VL-8B & 57.9 & 78.4 & 64.8 & 40.9 & 48.0 & 66.8 & 56.1 & 28.7 & 51.2 & 70.9 & 62.8 & 33.7 \\
            Qwen2.5-VL-72B & 52.1 & 71.6 & 52.9 & 42.2 & 50.2 & 68.0& 61.0& 35.0 & 40.1 & 62.8 & 44.8 & 18.6 \\
            \midrule
            \rowcolor{gray!10} \multicolumn{13}{l}{\textit{RS Specialized VLMs}} \\
            GeoChat & 17.6 & 30.1 & 4.3 & 0.1 & 7.8 & 11.4 & 1.1 & 0.0 & 12.6 & 20.9 & 2.9 & 0.0 \\
            SkySenseGPT & 23.4 & 46.8 & 11.3 & 0.7 & 12.7 & 23.6 & 6.7 & 0.2 & 2.5 & 4.7 & 1.7 & 0.0 \\
            VHM-7B &39.0 & 64.0 & 41.6 & 12.3 & 17.5 & 31.0 & 12.9 & 1.6 & 7.1 & 12.2 & 2.9 & 0.6 \\
            EarthDial-RGB & 47.6 & 76.9 & 54.7 & 17.3 & 37.4 & 59.9 & 40.5 & 15.4 & 19.1 & 33.1 & 18.0 & 3.5 \\
            EarthDial-MS & 0.0 & 0.0 & 0.0 & 0.0 & 0.1 & 0.2 & 0.0 & 0.0 & 0.0 & 0.0 & 0.0 & 0.0 \\
            GeoLLaVA-8K & 0.1 & 0.2 & 0.0 & 0.0 & 0.0 & 0.0 & 0.0 & 0.0 & 2.0 & 2.3 & 0.0 & 0.0 \\
            \bottomrule
        \end{tabular}
    }
    \vspace{2pt}
    \begin{flushleft}
    \footnotesize
    \textbf{Notes:} This table presents the baseline performance for object-level localization tasks. \textbf{A2.1 Visual Grounding} and \textbf{A2.2 Referring Expression Comprehension} focus on aligning textual queries with physical objects, while \textbf{B1.3 Functional Region Localization} requires high-level inference of zone utility. Metrics include \textbf{mIoU} and \textbf{Acc@X}, which denotes the accuracy when the IoU between the predicted bounding box and ground truth exceeds the threshold $X$. All numerical values are scaled by 100 for readability.
    \end{flushleft}
\end{table*}

\begin{table*}[!h]
    \caption{Full Evaluation of Detection Tasks: Delta Performance Metrics ($\Delta$).}
    \label{tab:det_delta}
    \centering
    \resizebox{\linewidth}{!}{
        \begin{tabular}{l | cccc | cccc | cccc}
            \toprule
            \multirow{2}{*}{\textbf{Method}} & \multicolumn{4}{c|}{\textbf{A2.1 Visual Grounding $\Delta$}} & \multicolumn{4}{c|}{\textbf{A2.2 Referring Expression Comprehension $\Delta$}} & \multicolumn{4}{c}{\textbf{B1.3 Functional Region Localization $\Delta$}} \\
            \cmidrule(lr){2-5} \cmidrule(lr){6-9} \cmidrule(lr){10-13}
            & mIoU & Acc@0.25 & Acc@0.5 & Acc@0.75 & mIoU & Acc@0.25 & Acc@0.5 & Acc@0.75 & mIoU & Acc@0.25 & Acc@0.5 & Acc@0.75 \\
            \midrule
            \rowcolor{gray!10} \multicolumn{13}{l}{\textit{General Close-source VLMs}} \\
            GPT-4o & 19.3 & 37.1 & 15.9 & 1.8 & 5.5 & 9.6 & 2.7 & 0.0 & 17.3 & 36.0 & 13.4 & 0.6 \\
            Claude-sonnet-4 & 8.0 & 17.0 & 10.1 & 1.4 & 2.5 & 7.1 & 1.3 & 0.0 & 21.5 & 40.7 & 27.9 & 5.8 \\
            Gemini-2.0-Flash & 19.0 & 38.1 & 25.1 & 3.8 & 12.7 & 25.4 & 10.2 & 0.4 & 20.6 & 41.9 & 18.0 & 3.5 \\
            \midrule
            \rowcolor{gray!10} \multicolumn{13}{l}{\textit{General Open-source VLMs}} \\
            GLM-4.6V & 46.1 & 59.6 & 57.4 & 41.2 & 51.0 & 63.5 & 62.8 & 47.9 & 47.8 & 68.6 & 61.6 & 37.8 \\
            InternVL3.5-38B & 45.9 & 54.5 & 64.0 & 47.3 & 8.6 & 12.3 & 12.7 & 7.6 & 34.1 & 50.6 & 38.4 & 21.5 \\
            InternVL3.5-8B & 42.4 & 59.0 & 52.6 & 29.9 & 15.1 & 20.7 & 18.9 & 8.7 & 28.1 & 44.2 & 29.1 & 12.2 \\
            InternVL3-8B & 14.0 & 24.6 & 17.0 & 6.7 & 10.2 & 19.2 & 3.1 & 0.2 & 13.2 & 26.2 & 2.9 & 0.6 \\
            Qwen3-VL-32B & 34.2 & 36.9 & 49.6 & 38.1 & 33.3 & 45.2 & 39.2 & 25.2 & 51.7 & 75.0 & 69.2 & 36.6 \\
            Qwen3-VL-8B & 46.2 & 59.3 & 63.3 & 40.9 & 40.8 & 57.9 & 55.0 & 28.7 & 47.1 & 65.7 & 61.6 & 33.7 \\
            Qwen2.5-VL-72B & 46.0 & 63.6 & 52.4 & 0.0 & 47.6 & 64.5 & 61.0 & 0.0 & 39.6 & 62.8 & 44.8 & 18.6 \\
            \midrule
            \rowcolor{gray!10} \multicolumn{13}{l}{\textit{RS Specialized VLMs}} \\
            GeoChat & 8.2 & 13.6 & 1.4 & 0.1 & 2.2 & 4.5 & 0.4 & 0.0 & 5.5 & 16.3 & 2.9 & 0.0 \\
            SkySenseGPT & 23.4 & 46.8 & 11.3 & 0.7 & 12.7 & 23.6 & 6.7 & 0.2 & 2.5 & 4.7 & 1.7 & 0.0 \\
            VHM-7B & 38.9 & 64.0 & 41.6 & 12.3 & 17.1 & 30.3 & 12.9 & 1.6 & 6.5 & 11.6 & 2.9 & 0.6 \\
            EarthDial-RGB & 44.8 & 74.3 & 54.3 & 17.3 & 35.0 & 57.9 & 39.9 & 15.4 & 18.0 & 32.0 & 17.4 & 3.5 \\
            EarthDial-MS & -0.1 & 0.0 & 0.0 & 0.0 & 0.1 & 0.2 & 0.0 & 0.0 & 0.0 & 0.0 & 0.0 & 0.0 \\
            GeoLLaVA-8K & 0.1 & 0.2 & 0.0 & 0.0 & 0.0 & 0.0 & 0.0 & 0.0 & 0.1 & 0.6 & -0.6 & 0.0 \\
            \bottomrule
        \end{tabular}
    }
    \vspace{2pt}
    \begin{flushleft}
    \footnotesize
    \textbf{Notes:} This table presents the \textbf{Visual Gain} ($\Delta$), which quantifies the improvement in performance when visual input is provided compared to the blind test (text-only) setting. It is calculated as $\Delta = Metric_{normal} - Metric_{blind}$. A higher positive $\Delta$ indicates that the model's localization and reasoning are strongly grounded in the actual image content. Conversely, low or near-zero $\Delta$ values suggest a heavy reliance on linguistic priors or dataset-specific spatial shortcuts. Negative values indicate cases where the presence of an image may introduce noise or lead to hallucinatory reasoning that degrades performance relative to the text-only baseline.
    \end{flushleft}
\end{table*}

\begin{table*}[!h]
    \caption{Full Evaluation of Generative Tasks: Base Performance Metrics. }
    \label{tab:gen_base}
    \centering
    \resizebox{\linewidth}{!}{
        \begin{tabular}{l | ccccc | ccccc | ccccc}
            \toprule
            \multirow{2}{*}{\textbf{Method}} & \multicolumn{5}{c|}{\textbf{A1.4 Image Captioning}} & \multicolumn{5}{c|}{\textbf{B2.1 Change Description}} & \multicolumn{5}{c}{\textbf{B2.3 Long-term Trend Reasoning}} \\
            \cmidrule(lr){2-6} \cmidrule(lr){7-11} \cmidrule(lr){12-16}
            & CIDEr & ROUGE-L & METEOR & BLEU-4 & BERT-Score & CIDEr & ROUGE-L & METEOR & BLEU-4 & BERT-Score & CIDEr & ROUGE-L & METEOR & BLEU-4 & BERT-Score \\
            \midrule
            \rowcolor{gray!10} \multicolumn{16}{l}{\textit{General Close-source VLMs}} \\
            GPT-4o & 151.9 & 20.6 & 22.5 & 33.8 & 87.4 & 113.3 & 22.2 & 23.4 & 34.7 & 88.5 & 68.9 & 16.0 & 13.9 & 25.8 & 84.3 \\
            Claude-sonnet-4 & 137.3 & 24.9 & 19.4 & 55.4 & 85.1 & 96.8 & 25.8 & 17.7 & 32.3 & 86.1 & 60.7 & 16.4 & 12.5 & 23.4 & 81.9 \\
            Gemini-2.0-Flash & 150.7 & 21.6 & 22.1 & 40.8 & 86.8 & 95.1 & 17.6 & 20.5 & 27.5 & 87.4 & 87.9 & 19.2 & 17.6 & 32.3 & 85.9 \\
            \midrule
            \rowcolor{gray!10} \multicolumn{16}{l}{\textit{General Open-source VLMs}}\\
            GLM-4.6V & 120.4 & 13.2 & 27.9 & 21.8 & 84.6 & 85.0 & 9.0 & 19.8 & 9.3 & 83.6 & 82.7 & 8.9 & 20.9 & 9.8 & 83.9 \\
            InternVL3.5-38B & 132.9 & 19.0 & 19.1 & 37.3 & 85.6 & 89.7 & 17.9 & 18.7 & 34.5 & 86.4 & 111.1 & 21.0 & 24.3 & 44.5 & 87.4 \\
            InternVL3.5-8B & 129.4 & 18.0 & 18.7 & 35.7 & 85.4 & 85.7 & 17.4 & 17.1 & 28.2 & 86.4 & 110.8 & 20.4 & 25.3 & 39.7 & 87.4 \\
            InternVL3-8B & 127.6 & 18.5 & 18.0 & 32.8 & 85.7 & 101.0 & 20.1 & 22.5 & 37.6 & 87.6 & 109.2 & 20.0 & 25.2 & 35.7 & 87.3 \\
            Qwen3-VL-32B & 131.8 & 18.1 & 25.3 & 52.8 & 85.0 & 92.2 & 15.1 & 26.8 & 22.1 & 85.5 & 86.7 & 14.5 & 24.6 & 24.4 & 84.7 \\
            Qwen3-VL-8B & 137.8 & 18.2 & 24.3 & 49.9 & 85.1 & 95.1 & 15.4 & 26.2 & 24.0 & 85.8 & 91.9 & 15.4 & 24.9 & 26.1 & 85.3 \\
            Qwen2.5-VL-72B & 80.0 & 10.9 & 14.4 & 29.1 & 81.7 & 105.1 & 18.9 & 25.7 & 36.1 & 87.2 & 100.7 & 17.1 & 26.3 & 29.9 & 86.1 \\
            \midrule
            \rowcolor{gray!10} \multicolumn{16}{l}{\textit{RS Specialized VLMs}} \\
            GeoChat & 138.1 & 20.8 & 17.6 & 25.2 & 86.0 & 102.4 & 19.7 & 22.4 & 39.5 & 87.1 & 99.0 & 18.7 & 22.5 & 40.7 & 86.6 \\
            SkySenseGPT & 125.6 & 19.1 & 15.8 & 22.4 & 85.6 & 95.6 & 18.2 & 21.2 & 37.8 & 86.7 & 99.0 & 19.1 & 21.7 & 41.9 & 86.7 \\
            VHM-7B & 113.0 & 17.6 & 12.6 & 12.4 & 85.5 & 93.8 & 19.8 & 17.8 & 34.9 & 87.1 & 100.6 & 19.7 & 19.6 & 44.2 & 87.0 \\
            EarthDial-RGB & 72.7 & 11.4 & 7.2 & 6.3 & 82.5 & 68.1 & 15.3 & 10.0 & 9.6 & 85.8 & 74.6 & 15.7 & 12.8 & 24.6 & 85.7 \\
            EarthDial-MS & 66.3 & 7.4 & 3.5 & 2.5 & 80.7 & 71.5 & 15.3 & 8.3 & 2.5 & 84.5 & 59.5 & 12.1 & 6.2 & 1.5 & 82.6 \\
            \bottomrule
        \end{tabular}
    }
    \vspace{2pt}
    \begin{flushleft}
    \footnotesize
    \textbf{Notes:} This table summarizes the performance on open-ended text generation tasks. \textbf{A1.4 Image Captioning} assesses descriptive fluency, \textbf{B2.1 Change Description} requires identifying differences between bi-temporal images, and \textbf{B2.3 Long-term Trend Reasoning} evaluates the interpretation of multi-temporal sequences. We utilize a comprehensive suite of metrics: \textbf{CIDEr} (Consensus-based Image Description Evaluation) measures the agreement with human consensus; \textbf{ROUGE-L} and \textbf{METEOR} assess recall and alignment; \textbf{BLEU-4} measures n-gram precision; and \textbf{BERT-Score} evaluates semantic similarity using contextual embeddings. Note that \textbf{CIDEr} values are scaled by 1000, while all other metrics are scaled by 100 to ensure clarity in comparison.
    \end{flushleft}
\end{table*}

\begin{table*}[!h]
    \caption{Full Evaluation of Generative Tasks: Delta Performance Metrics ($\Delta$). }
    \label{tab:gen_delta}
    \centering
    \resizebox{\linewidth}{!}{
        \begin{tabular}{l | ccccc | ccccc | ccccc}
            \toprule
            \multirow{2}{*}{\textbf{Method}} & \multicolumn{5}{c|}{\textbf{A1.4 Image Captioning $\Delta$}} & \multicolumn{5}{c|}{\textbf{B2.1 Change Description $\Delta$}} & \multicolumn{5}{c}{\textbf{B2.3 Long-term Trend Reasoning $\Delta$}} \\
            \cmidrule(lr){2-6} \cmidrule(lr){7-11} \cmidrule(lr){12-16}
            & CIDEr & ROUGE-L & METEOR & BLEU-4 & BERT-Score & CIDEr & ROUGE-L & METEOR & BLEU-4 & BERT-Score & CIDEr & ROUGE-L & METEOR & BLEU-4 & BERT-Score \\
            \midrule
            \rowcolor{gray!10} \multicolumn{16}{l}{\textit{General Close-source VLMs}} \\
            GPT-4o & 100.4 & 13.5 & 13.7 & 21.1 & 6.2 & 85.0 & 17.2 & 15.2 & 27.4 & 5.5 & 32.1 & 9.4 & 4.2 & 14.5 & 1.2 \\
            Claude-sonnet-4 & 28.5 & 5.5 & 2.7 & 4.9 & 1.3 & 17.9 & 5.4 & 1.8 & 0.6 & 0.9 & -17.7 & -5.7 & -3.1 & -12.6 & -2.8 \\
            Gemini-2.0-Flash & 67.2 & 10.3 & 8.9 & 16.7 & 4.5 & 38.3 & 8.7 & 7.7 & 16.0 & 3.5 & 18.8 & 4.4 & 3.6 & -5.6 & 2.1 \\
            \midrule
            \rowcolor{gray!10} \multicolumn{16}{l}{\textit{General Open-source VLMs}}\\
            GLM-4.6V & 31.0 & 0.6 & 6.9 & -8.0 & 2.3 & 24.4 & 1.7 & 4.9 & 0.2 & 2.2 & 16.0 & 1.4 & 2.9 & 0.4 & 1.7 \\
            InternVL3.5-38B & 30.0 & 3.6 & 2.5 & 0.5 & 2.7 & 20.7 & 3.7 & 3.4 & 10.4 & 1.7 & 41.4 & 8.6 & 5.9 & 20.6 & 4.2 \\
            InternVL3.5-8B & 22.0 & 2.2 & 0.7 & -3.8 & 2.3 & 19.5 & 3.6 & 2.4 & 6.9 & 1.7 & 45.5 & 9.0 & 7.3 & 17.7 & 4.8 \\
            InternVL3-8B & 63.2 & 6.9 & 8.7 & 17.4 & 2.4 & 28.7 & 3.7 & 7.8 & 9.2 & 1.7 & 35.5 & 4.4 & 9.1 & 6.6 & 1.7 \\
            Qwen3-VL-32B & 57.2 & 5.9 & 10.9 & 17.5 & 3.0 & 42.2 & 4.3 & 11.8 & -1.8 & 3.1 & 32.5 & 3.3 & 8.3 & -4.5 & 2.6 \\
            Qwen3-VL-8B & 61.1 & 5.6 & 9.9 & 13.0 & 3.0 & 46.1 & 4.5 & 11.8 & -1.7 & 3.7 & 39.8 & 4.2 & 9.9 & -6.7 & 3.1 \\
            Qwen2.5-VL-72B & 21.8 & 2.5 & 3.2 & 6.1 & 1.4 & 60.1 & 9.8 & 14.2 & 20.5 & 4.9 & 43.5 & 6.6 & 11.1 & 12.9 & 3.3 \\
            \midrule
            \rowcolor{gray!10} \multicolumn{16}{l}{\textit{RS Specialized VLMs}} \\
            GeoChat & 16.5 & 2.0 & 2.5 & 4.9 & 1.3 & 12.1 & 2.7 & 3.6 & 5.4 & 1.7 & 9.0 & 1.9 & 2.4 & 2.7 & 1.2 \\
            SkySenseGPT & 2.2 & 0.4 & 0.9 & 3.3 & 0.1 & 4.0 & 1.2 & 1.0 & 2.5 & 0.6 & 9.1 & 2.5 & 1.0 & 7.3 & 0.9 \\
            VHM-7B & 0.9 & -0.5 & -0.4 & -2.6 & 0.0 & 2.7 & 0.8 & 0.4 & -1.4 & 0.4 & 4.9 & 0.2 & 0.8 & 2.7 & 0.3 \\
            EarthDial-RGB & 7.9 & 1.4 & 1.2 & 2.6 & -0.3 & 11.2 & 2.6 & 1.9 & 2.1 & 0.8 & 23.1 & 4.4 & 4.5 & 11.5 & 1.9 \\
            EarthDial-MS & -4.0 & 1.0 & 0.7 & 1.6 & -0.5 & -0.4 & 0.1 & 0.3 & 0.9 & -0.5 & 3.0 & 0.6 & 1.1 & 0.9 & -0.5 \\
            GeoLLaVA-8K & 0.0 & 0.0 & 0.0 & 0.0 & 0.0 & 0.0 & 0.0 & 0.0 & 0.0 & 0.0 & 0.0 & 0.0 & 0.0 & 0.0 & 0.0 \\
            \bottomrule
        \end{tabular}
    }
    \vspace{2pt}
    \begin{flushleft}
    \footnotesize
    \textbf{Notes:} This table illustrates the \textbf{Visual Gain} ($\Delta$) for generative tasks, defined as the metric difference between the normal multimodal input and the text-only blind test. \textbf{C-$\Delta$}, \textbf{R-$\Delta$}, \textbf{M-$\Delta$}, \textbf{B4-$\Delta$}, and \textbf{B-S-$\Delta$} represent the gain in CIDEr, ROUGE-L, METEOR, BLEU-4, and BERT-Score, respectively. A positive $\Delta$ indicates that the model successfully utilizes visual evidence to refine its descriptions. Notably, low $\Delta$ values in complex reasoning tasks reveal that many models rely heavily on temporal templates or linguistic logical chains rather than actual pixel-level changes. Negative values suggest that misleading visual noise or poor cross-modal alignment may cause the model to generate descriptions less accurate than those produced by pure linguistic inference.
    \end{flushleft}
\end{table*}

\end{document}